\documentclass[10pt,twocolumn,letterpaper]{article}

\usepackage{iccv}              
\usepackage{dsfont}
\usepackage{amsthm}
\usepackage{algorithm}
\usepackage{algpseudocode}
\usepackage{comment}
\newtheorem{definition}{Definition}
\newtheorem{proposition}{Proposition}
\newtheorem{theorem}{Theorem}

\definecolor{iccvblue}{rgb}{0.21,0.49,0.74}
\usepackage[pagebackref,breaklinks,colorlinks,allcolors=iccvblue]{hyperref}

\def\paperID{8972} 
\def\confName{ICCV}
\def\confYear{2025}

\title{FedXDS: Leveraging Model Attribution Methods to counteract Data Heterogeneity in Federated Learning}

\author{
Maximilian Andreas Hoefler$^{1}$ \thanks{Correspondence to maximilian.andreas.hoefler@hhi.fraunhofer.de}  \quad
Karsten Mueller$^{1}$ \quad
Wojciech Samek$^{1,2,3}$ \\
$^{1}$Fraunhofer Heinrich Hertz Institute, Berlin, Germany \\
$^{2}$Technical University of Berlin, Berlin, Germany \\
$^{3}$Berlin Institute for the Foundations of Learning and Data (BIFOLD) 
}
\begin{document}
\maketitle
\begin{abstract}
Explainable AI (XAI) methods have demonstrated significant success in recent years at identifying relevant features in input data that drive deep learning model decisions, enhancing interpretability for users. However, the potential of XAI beyond providing model transparency has remained largely unexplored in adjacent machine learning domains. In this paper, we show for the first time how XAI can be utilized in the context of federated learning. Specifically, while federated learning enables collaborative model training without raw data sharing, it suffers from performance degradation when client data distributions exhibit statistical heterogeneity. We introduce FedXDS (Federated Learning via XAI-guided Data Sharing), the first approach to utilize feature attribution techniques to identify precisely which data elements should be selectively shared between clients to mitigate heterogeneity. By employing propagation-based attribution, our method identifies task-relevant features through a single backward pass, enabling selective data sharing that aligns client contributions. To protect sensitive information, we incorporate metric privacy techniques that provide formal privacy guarantees while preserving utility. Experimental results demonstrate that our approach consistently achieves higher accuracy and faster convergence compared to existing methods across varying client numbers and heterogeneity settings. We provide theoretical privacy guarantees and empirically demonstrate robustness against both membership inference and feature inversion attacks. Code is available at \url{https://github.com/MaxH1996/FedXDS}.
\end{abstract}

\section{Introduction}
Federated learning (FL) \cite{FedAvg} has received significant attention in recent years, proving to be an effective method for collaborative distributed machine learning while keeping local datasets private. However, the practical deployment of federated learning systems faces fundamental challenges regarding data heterogeneity and privacy \cite{kairouz2021advances,zhao2018federated}. 

Data heterogeneity, characterized by non-IID data across clients, leads to contradicting model updates and significantly degraded performance \cite{SCAFFOLD, FLheterogeneous}. Current approaches address this challenge by aligning client distributions using proximal optimization terms \cite{FedProx} and mitigating client drift \cite{SCAFFOLD, mikkelschmidt}, while others focus on flattening the loss landscape to facilitate better model aggregation \cite{LocalLearningMatters, FedSAM}. In addition, recent works have shown that sharing even a small portion of client data globally can significantly improve local model generalization \cite{data_sharing_FL}. Such shared data introduces centralization, giving clients access to a more homogeneous distribution, which reduces update divergence and mitigates heterogeneity \cite{data_sharing_FL}.

However, sharing raw data is not feasible due to privacy constraints. Introducing differential privacy \cite{dwork2006differential} by adding noise to the shared data offers a potential solution, however this can cause considerable performance loss. Alternatively, instead of sharing raw data, abstract feature representations or partial data could be released using generator-type approaches \cite{FedFTG, FedGen, FedFed}. However, the obtained features are not as potent as raw data in mitigating statistical heterogeneity, and can also be prone to privacy leaks and additionally introduce significant computational overhead due to generators. Hence, the challenge arises of \textit{ how to retain the performance improvement from data sharing while simultaneously guaranteeing privacy}.

To address this challenge we propose FedXDS (Federated Learning via XAI-guided Data Sharing) which entails a novel approach wherein we leverage methods from the field of eXplainable AI (XAI) \cite{XAI_review_paper} enabling performance enhancing data sharing while preserving privacy.

Specifically, we leverage propagation-based attribution methods, which have been shown to reliably identify input features that consistently contribute to model predictions \cite{XAI_review_paper, xai-benchmark}. Moreover, attribution maps yield a pixel-wise relevance score which highlights regions in the input space which focus on semantically meaningful features, while suppressing spurious correlations and background noise. In our setup local client data is filtered through the attribution mechanism, i.e., we only share the input features which the attribution map deems most relevant for model predictions. This retains the information needed for model generalization while discarding irrelevant information.

We show that our method, when combined with differential-metric privacy techniques \cite{dwork2006differential, d_x-privacy}, offers stronger privacy guarantees compared to applying similar privacy mechanisms directly to raw input features. Specifically, by discarding task-irrelevant information we effectively reduce the dimensionality of the raw features. This allows us to selectively protect only the task-relevant regions rather than uniformly protecting all input pixels, and thus obtain better utility. 

In addition, propagation-based attributions can be obtained via a single backward pass over the local dataset, which only needs to be performed once in the entire FL process.

\noindent Our main contributions can be summarized as:
\begin{itemize}
\item A novel federated learning algorithm leveraging propagation-based attributions to address the challenge of retaining performance from data sharing with privacy guarantees in heterogeneous environments.
\item A privacy-preserving mechanism that utilizes attribution-guided dimensionality reduction to achieve metric differential privacy. We perform theoretical and empirical privacy evaluations on our feature sharing approach through membership inference attacks and feature inversion, demonstrating that we can attain strong privacy guarantees.
\item Extensive experiments on image datasets benchmarking  our method against current state of the art, demonstrating superior performance in terms of accuracy and convergence.
\end{itemize}
\section{Preliminaries}
\subsection{Federated Learning}
Federated Learning \cite{FedAvg} is a distributed machine learning paradigm where multiple clients collaborate to train a model under the orchestration of a central server, without sharing their raw data. Let $K$ be the number of clients, each with a local dataset $\mathcal{D}_k = \{(x_i^k, y_i^k)\}_{i=1}^{n_k}$, where $x_i^k \in \mathcal{X}$ is an input sample and $y_i^k \in \mathcal{Y}$ is its corresponding label. The goal is to learn a global model $f_\theta: \mathcal{X} \rightarrow \mathcal{Y}$ parameterized by $\theta$, which minimizes the empirical risk:

\begin{equation}
    \min_\theta \frac{1}{N} \sum_{k=1}^K \sum_{i=1}^{n_k} \mathcal{L}(f_\theta(x_i^k), y_i^k)
\end{equation}

where $N = \sum_{k=1}^K n_k$ is the total number of samples across all clients, and $\mathcal{L}$ is a suitable loss function.

The FedAvg algorithm \cite{FedAvg} iteratively aggregates model updates from selected clients $S_t$, where each client $k$ performs local updates $\theta_k^t = \theta^{t-1} - \eta \nabla \mathcal{L}_k(\theta^{t-1})$ before the server averages these updates as $\theta^t = \frac{1}{|S_t|} \sum_{k \in S_t} \theta_k^t$.
\subsection{Differential Privacy in Metric Spaces}

Standard differential privacy (DP) \cite{dwork2006differential} provides guarantees against membership inference by bounding the output distribution change when a single dataset element is modified. For continuous data such as image embeddings, classical adjacency notions are restrictive, making \emph{metric differential privacy} \cite{d_x-privacy} more appropriate:

\begin{definition}[Metric Privacy] \cite{d_x-privacy}
\label{def:metric-privacy}
Let $(X, d_X)$ be a metric space and $Z$ a set of possible outputs. A randomized mechanism $A \colon X \to \Delta(Z)$ is \emph{$(\varepsilon,\delta)$-metric private} if for all $x,x'\in X$ and every measurable set $U\subseteq Z$:
\[
\Pr[A(x)\in U] \;\le\; \exp\Bigl(\varepsilon\,d_X(x,x')\Bigr)\,\Pr[A(x')\in U] \;+\; \delta.
\]
\end{definition}

This definition reduces to standard DP when $d_X(x,x')=1$ for all adjacent datasets, but naturally captures similarity in continuous spaces. In our work, we define $(X, d_X)$ where $X = \mathbb{R}^d$ represents image data and $d_X(x,x') =||x-x'||_2$ is the $\ell_2$ distance between inputs.

To guarantee privacy, we calibrate noise based on the sensitivity of a query function $f\colon X \to \mathbb{R}^m$, which measures output changes relative to input changes. In our work we choose the following sensitivity measure as used in \cite{CI-lipschitz, test_lipschitz, restricted_privacy}:

\begin{definition}[Sensitivity]
\label{def:global-sensitivity}
For a function $f\colon X \to \mathbb{R}^m$, the sensitivity is:
\[
\Delta_f = \max_{x,x'\in X} \frac{\|f(x)-f(x')\|}{||x-x'||}.
\]
\end{definition}

This definition captures the largest possible output difference relative to input difference, directly controlling how much noise is needed for privacy. This notion of sensitivity offers several advantages, which we discuss in the supplementary material. Importantly, when $\Delta_f$ is small, less noise is required to achieve the same privacy guarantee, preserving more signal and improving utility—particularly valuable for high-dimensional data like images.

For privacy preservation, we employ the Gaussian mechanism, which adds noise to the output of our query function proportional to the sensitivity $\Delta_f$:

\begin{theorem}[Gaussian Mechanism] \cite{dwork2006differential, d_x-privacy}
\label{thm:gauss}
For a function $f: \mathcal{D} \rightarrow \mathbb{R}^d$ with sensitivity $\Delta_f$, adding Gaussian noise with scale $\sigma$ satisfying:
\begin{equation}
   \sigma \geq \frac{\Delta_f \sqrt{2\log(1.25/\delta)}}{\varepsilon}
\end{equation}
ensures $f(x) + \mathcal{N}(0, \sigma^2 I)$ is $(\varepsilon, \delta)$-metric private.
\end{theorem}

This mechanism forms the foundation of our privacy-preserving feature sharing approach, allowing us to bound and quantify the privacy guarantees of our federated learning framework.
\subsection{Neural Network Attribution Methods}
\label{sec:attribution_methods}
We consider several gradient-based attribution techniques in our work. The simplest being Gradient × Input \cite{gradient}, which computes attributions through element-wise multiplication of input and gradient:
\begin{equation}
    \mathcal{A}_{\text{grad}}(f_\theta, \mathbf{x}) = \mathbf{x} \odot \frac{\partial f_\theta}{\partial \mathbf{x}}
\end{equation}
Integrated Gradients (IG) \cite{integrated_grad} addresses gradient saturation by accumulating gradients along a path from baseline $\mathbf{x}'$ to input:
\begin{equation}
    \mathcal{A}_{\text{IG}}(f_\theta, \mathbf{x}) = (\mathbf{x} - \mathbf{x}') \odot \int_0^1 \frac{\partial f_\theta(\mathbf{x}' + \alpha(\mathbf{x} - \mathbf{x}'))}{\partial \mathbf{x}} d\alpha
\end{equation}
SmoothGrad \cite{smoothgrad} reduces attribution noise by averaging gradients over perturbed inputs:
\begin{equation}
    \mathcal{A}_{\text{smooth}}(f_\theta, \mathbf{x}) = \frac{1}{N}\sum_{i=1}^N \frac{\partial f_\theta}{\partial (\mathbf{x} + \epsilon)}
\end{equation}
Lastly we use Layer-wise Relevance Propagation (LRP) with the $\epsilon$-rule according to \cite{bach2015pixel}. This recursively propagates relevance scores $R^{(l)}$ from layer $l$ to $l-1$ using:
\begin{equation}
   R_j = \sum \frac{a_jw_{jk}}{\epsilon + \sum_{0,j} a_jw_{jk}} R_k
\end{equation}
where $w_{jk}$ are layer weights, $a_j$ are layer activations, and $\epsilon$ is a small stabilizing term.
\section{Related Work}
\label{sec:related_work}

Federated Learning was introduced with FedAvg~\cite{FedAvg}, enabling collaborative model training without sharing raw data, but suffers under statistical heterogeneity. Regularization-based methods mitigate client drift by constraining local updates, such as FedProx~\cite{FedProx}, SCAFFOLD~\cite{SCAFFOLD}, and FedDyn~\cite{FedDyn}, with extensions like FedBN~\cite{FedBN}, MOON~\cite{MOON}, FedNova~\cite{FedNova}, FedSAM~\cite{FedSAM}, and FedDISCO~\cite{Feddisco} targeting specific aspects like feature shifts or loss geometry. A more direct solution is data or knowledge sharing: FedDF and FedAux~\cite{FedDF, FedAux} distill client knowledge using a public dataset; FedGen~\cite{FedGen} synthesizes class-conditional features for distribution alignment; FedFTG~\cite{FedFTG} generates pseudo-data for feature-level transfer; and FedFed~\cite{FedFed} applies VAE-based feature distillation. While effective, these methods often introduce significant computational and privacy costs due to data synthesis and sharing. In contrast, FedXDS shares compact attribution-based feature subsets derived via XAI, avoiding generators while enhancing privacy, efficiency, and performance in heterogeneous settings. Wo provide an extense related works in the supplementary material.

\section{FedXDS Approach}
\begin{figure}[!t]
\centering
\includegraphics[width=3.0in]{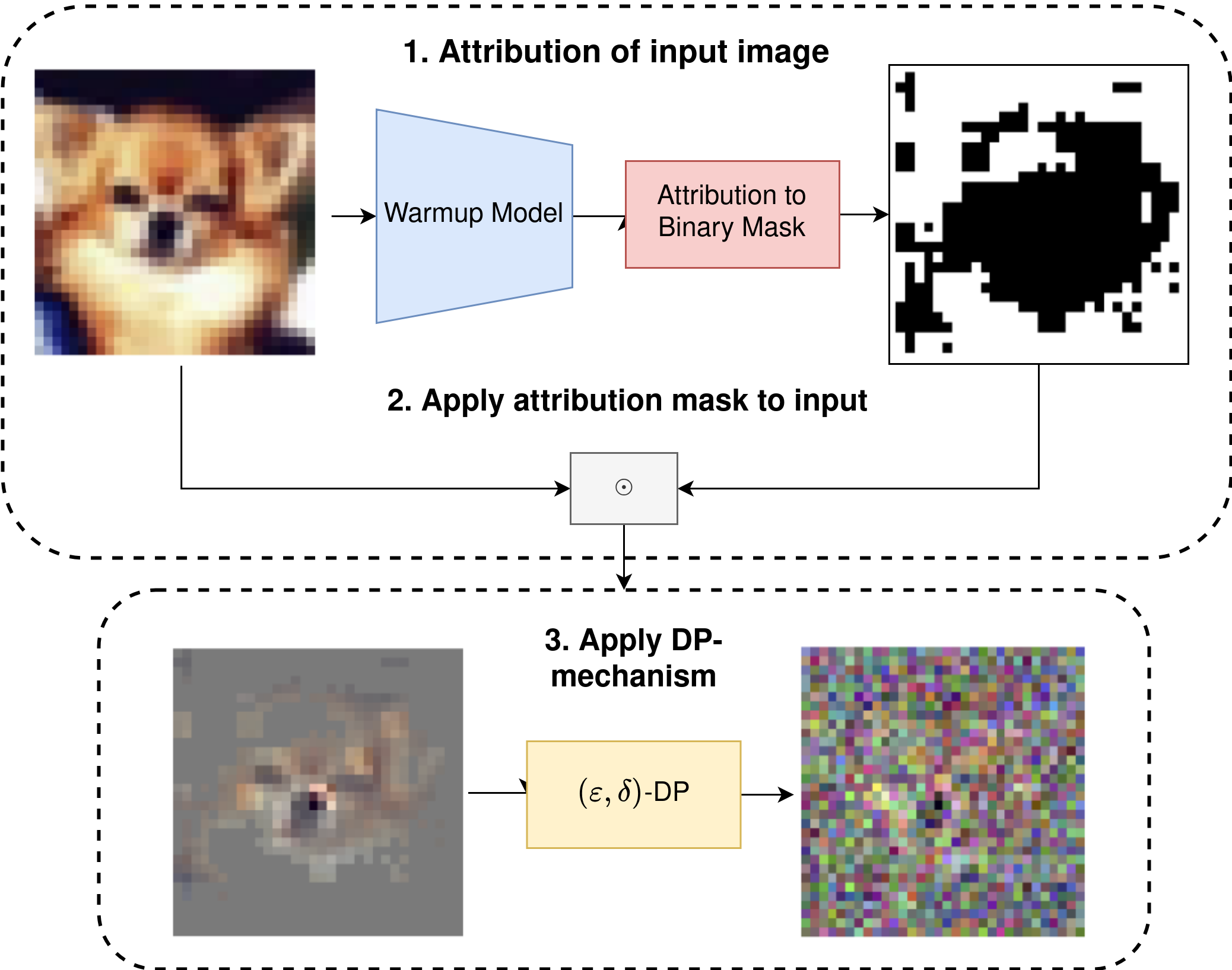}
\caption{Illustration of the DP attribution-guided feature extraction. The process consists of four main steps: (1) Computing attribution scores using a warmup model to identify important features, (2) Creating and applying a binary mask based on the attributions, (3) Adding Gaussian noise for metric privacy.}
\label{fig:FedXDS_dp_image_generation}
\end{figure}

\begin{algorithm}[t]
\caption{Attribution-Guided Private Representation Extraction.}
\label{alg:private_rep}
\begin{algorithmic}[1]
\Require $D_k$: client dataset, $f_\theta$: model, $s$: sparsity level, $(\varepsilon,\delta)$: privacy params, $S$: sensitivity, $\theta_{\text{warmup}}$
\Ensure Private dataset $D_k^p$

\For{$\mathbf{x}_i^k \in D_k$}
    \State $\mathbf{h} \gets \mathcal{A}(f_{\theta_{\text{warmup}}}, \mathbf{x}_i^k)$; $h^{(s)} \gets$ s-th largest in $\mathbf{h}$
    \State $\mathbf{m} \gets [\mathbf{h}[i] \geq h^{(s)}]$; $\mathbf{x}_m \gets \mathbf{x}_i^k \odot \mathbf{m}$ 
    \State $\tilde{\mathbf{x}}_i^k \gets \mathbf{x}_m + \mathcal{N}(0, \sigma^2)$ \Comment{$\sigma$ from $(\varepsilon,\delta)$}
    \State Add $(\tilde{\mathbf{x}}_i^k, y_i^k)$ to $D_k^p$
\EndFor
\State \Return $D_k^p$
\end{algorithmic}
\end{algorithm}

\begin{algorithm}[t]
\caption{Privacy-Preserving Federated Training}
\label{alg:federated_training}
\begin{algorithmic}[1]
\Require $\{D_k\}_{k=1}^K$: client datasets, $\lambda$: knowledge weight, $T$: rounds, $E$: local epochs, $\eta$: learning rate, $R_{\textbf{warmup}}$

\State Initialize $\theta_{\text{warmup}}$ using $R_{\textbf{warmup}}$ rounds of FedAvg
\State Obtain $D_k^p$ using Algorithm \ref{alg:private_rep} for all $k \in [K]$ in parallel
\State Server: $\mathcal{D}_g \gets \bigcup_{k=1}^K D_k^p$

\For{round $t = 1$ to $T$}
    \State Select client subset $S_t \subseteq [K]$
    \For{$k \in S_t$ \textbf{in parallel}}
        \State $\theta_k^t \gets \theta^{t-1}$ \Comment{Download global model}
        \For{epoch $e = 1$ to $E$}
            \For{batch $B$ from $D_k$, $B_g$ from $\mathcal{D}_g$}
                \State $\mathcal{L}_k^t \gets \frac{1}{|B_k|}\sum_{(\mathbf{x}, y) \in B_k} \ell(f_{\theta_k^t}(\mathbf{x}), y) + \lambda \frac{1}{|B_g|}\sum_{(\mathbf{x}_g, y_g) \in B_g} \ell(f_{\theta_k^t}(\mathbf{x}_g), y_g)$
                \State $\theta_k^t \gets \theta_k^t - \eta \nabla \mathcal{L}$
            \EndFor
        \EndFor
    \EndFor
    \State Server: $\theta^t \gets \frac{1}{|S_t|} \sum_{k \in S_t} \theta_k^t$ \Comment{FedAvg}
\EndFor
\end{algorithmic}
\end{algorithm}
 Consider a federated learning system with $N$ clients, where each client $k$ holds local data $\mathcal{D}_k$. In our setting all clients share the same neural network architecture $f_\theta$. Our goal is to learn a global model $\Theta_G$ which generalizes well on all client datasets.

\subsection{Attribution-Guided Feature Selection}
\label{subsec:feature_selection}
The first stage of our method identifies discriminative features using a pre-trained warmup model, with parameters $\theta_{\text{warmup}}$. This is achieved by training with FedAvg for $R_{warmup}$ rounds, before sharing representations. Then, for each input sample $\mathbf{x} \in \mathbb{R}^{H\times W\times 3}$, where $H$ and $W$ are the dimensions of the image with three color channels, in the client dataset $\mathcal{D}_k$, we compute relevances using an attribution method $\mathcal{A}$ from Section~\ref{sec:attribution_methods} :
\begin{equation}
    \mathbf{h} = \mathcal{A}(f_\theta, \mathbf{x})
\end{equation}
where $\mathbf{h} \in \mathbb{R}^{H\times W}$ is the pixel-wise importance score of an input image $\mathbf{x}$. These attribution scores quantify the importance of each input feature to the model's prediction. To focus on the most relevant features, we create a binary mask $\mathbf{m}$, as shown in \autoref{eq:binary_mask}, by retaining features whose attribution scores exceed a threshold determined by the desired sparsity level $s$:

\begin{equation}
    \mathbf{m}[i] = \begin{cases} 
        1 & \text{if } \mathbf{h}[i] \geq h^{(s)} \\
        0 & \text{otherwise}
    \end{cases}
    \label{eq:binary_mask}
\end{equation}

where $h^{(s)}$ represents the $s$-th largest value in the attribution scores. This mask identifies the subset of features that are most crucial for the model's decision-making process.

We define our feature selection function $f_A: \mathbb{R}^{H\times W\times 3} \rightarrow \mathbb{R}^{H\times W\times 3}$ as the application of the attribution mask $\mathbf{m}$ to each image $\mathbf{x} \in \mathcal{D}_k$ through:
\begin{equation}
    f_A(\mathbf{x}) = \mathbf{x} \odot \mathbf{m}
    \label{eq:feature_selection}
\end{equation}

This operation emphasizes the most informative parts of the input while suppressing less relevant details. However, directly sharing $f_A(\mathbf{x})$ would still pose privacy risks. Therefore, we design a privacy mechanism $\mathcal{M}$ that adds calibrated noise to these masked features (Section~ref{subsec:privacy}):

\begin{equation}
    \mathcal{M}(\mathbf{x}) = f_A(\mathbf{x}) + \mathcal{N}(0, \sigma^2\mathbf{I})
    \label{eq:privacy_mechanism}
\end{equation}

where $\sigma$ is determined based on privacy requirements. The resulting privacy-protected features, along with their corresponding labels, are then aggregated into a new dataset $\mathcal{D}_k^p = \{(\mathcal{M}(\mathbf{x}), y) | (\mathbf{x}, y) \in \mathcal{D}_k\}$ and shared with the server. The full data generated process is outlined in \autoref{alg:private_rep} and visualized in \autoref{fig:FedXDS_dp_image_generation}.

\subsection{Local Training using Shared Data}
After the privacy-protected features have been obtained, the server aggregates private representations from all clients into a global dataset $\mathcal{D}_g = \bigcup_{k=1}^N \mathcal{D}_k^p$, and sends this dataset to each client. The clients then optimize a composite objective that balances local task performance with knowledge from the global private dataset:
\begin{equation}
\begin{split}
    \min_\theta & \big[\ \mathbb{E}_{(\mathbf{x},y) \sim \mathcal{D}_k}[\ell(f_\theta(\mathbf{x}), y)] \\
    &+ \lambda \mathbb{E}_{(\mathbf{x},y) \sim \mathcal{D}_g}[\ell(f_\theta(\mathbf{x}), y)] \big]
\end{split}
\label{eq:fedxds_learning_objective}
\end{equation}
The hyperparameter $\lambda$ controls the trade-off between local specialization and global knowledge integration. This formulation allows clients to benefit from collective learning while maintaining their own task-specific performance and ensuring privacy. The procedure is outlined in \autoref{alg:federated_training}.

\subsection{Privacy Discussion}
\label{subsec:privacy}
We build on the insight that sharing features derived from raw client data can mitigate data heterogeneity in federated learning \cite{data_sharing_FL}. Nevertheless, directly sharing these features poses significant privacy risks. A straightforward approach would be to apply differential privacy (DP) to the raw data itself; however, the high dimensionality of image data forces the addition of very large noise levels, thereby significantly degrading utility.

To address this challenge, we propose a two-step approach. First, we use an attribution-guided masking strategy (Section~\ref{subsec:feature_selection}) to identify and retain only task-relevant pixels. Second, we add Gaussian noise (\autoref{thm:gauss}) to the resulting masked features to achieve the desired $\varepsilon$-metric privacy (\autoref{def:metric-privacy}, \autoref{def:global-sensitivity}).

This design offers two main advantages. First, by sparsifying input features that do not contribute to classification predictions, we reduce the overall sensitivity of the data. According to \autoref{thm:gauss}, lower sensitivity requires less noise to meet the same privacy budget $\varepsilon$, thus improving utility. Second, in contrast to random or magnitude-based sparsification that may unintentionally discard valuable information, our attribution-guided approach specifically preserves the pixels most critical for the learning task while eliminating irrelevant ones.

The primary goal of our privacy mechanism is to protect against membership inference attacks (MIA) and feature inversion, which we empirically validate in \autoref{sec:privacy_analysis}. In the following section, we formally demonstrate how our strategy guarantees metric privacy, while a comprehensive discussion of privacy assumptions appears in the appendix.

\subsubsection{Sensitivity Analysis of Attribution-Based Masking}

A critical component of ensuring privacy is understanding the sensitivity of our feature selection function $f(\mathbf{x})$ defined in \autoref{eq:feature_selection}. From \autoref{thm:gauss}, the noise required to attain a certain privacy level $\varepsilon$ depends on the sensitivity $\Delta_f$. Our goal is to minimize this sensitivity before applying the privacy mechanism $\mathcal{M}$.

For our feature selection function $f_A(\mathbf{x}) = \mathbf{x} \odot \mathbf{m}$, we can derive a strict bound on sensitivity. Since each coordinate of the mask satisfies $\mathbf{m}[i] \in \{0,1\}$, for any two inputs $\mathbf{x}, \mathbf{x}' \in \mathbb{R}^{H\times W\times 3}$, we have, in the worst case:
\[
|\mathbf{m}[i] \mathbf{x}[i] - \mathbf{m}[i] \mathbf{x}'[i]| \leq |\mathbf{x}[i] - \mathbf{x}'[i]|
\]

Thus, in aggregate:
\[
\|f_A(\mathbf{x}) - f_A(\mathbf{x}')\|_2 \leq \|\mathbf{x} - \mathbf{x}'\|_2
\]

This shows that our masking operation is non-expansive with respect to the $\ell_2$ norm. Rearranging the equation:
\begin{equation}
   \Delta_f =  \frac{\|f_A(\mathbf{x}) - f_A(\mathbf{x}')\|_2}{\|\mathbf{x} - \mathbf{x}'\|_2} \leq 1
\end{equation}
where $\Delta_f$ is the sensitivity we defined in the preliminaries in \autoref{def:global-sensitivity}.
Without our attribution-based masking, the sensitivity would be unbounded for arbitrary transformations. Moreover, in the worst case $f_A$ is the identity, i.e. $f_A(x) = x$, meaning that the attribution method deems all pixels important. This implies that our method would reduce to the naive approach of adding noise to the raw features. In practice, this rarely occurs, since this would already require a highly sparsified image. Nevertheless, our approach safeguards against this worst-case scenario. More discussion can be found in the supplementary material. 


\subsubsection{Privacy Mechanism and Guarantees}

Based on the sensitivity bound established above, we design our privacy mechanism $\mathcal{M}$ (Equation \ref{eq:privacy_mechanism}) by adding Gaussian noise calibrated to this reduced sensitivity:
\[
\mathcal{M}(\mathbf{x}) = f_A(\mathbf{x}) + \mathcal{N}(0, \sigma^2\mathbf{I})
\]

\noindent where $\sigma$ is chosen according to \autoref{thm:gauss} for an $(\varepsilon, \delta)$-level of privacy given sensitivity $\Delta_f \leq 1$. This ensures that for all input images in the client dataset, $\mathbf{x},\mathbf{x}' \in D_k$, our method satisfies $(\varepsilon,\delta)$-metric privacy. 

Our theoretical privacy guarantees are empirically validated through membership inference and feature inversion attacks Section~\ref{sec:privacy_analysis}, where attribution-masked features consistently show stronger protection than unmasked features under equivalent privacy budgets.

\section{Experimental}
\subsection{Experimental Setup}
We evaluate FedXDS on standard image classification benchmarks (CIFAR-10, CIFAR-100, Tiny-ImageNet) and real-world federated datasets from the LEAF framework \cite{leaf} (CelebA, FEMNIST) that naturally exhibit heterogeneous distributions. For the standard benchmarks, we simulate statistical heterogeneity using Dirichlet-based partitioning with concentration parameters \(\alpha \in \{0.05,0.1\}\), where smaller values indicate greater heterogeneity. We conduct experiments with both 10 and 100 clients, setting participation rates to 0.5 and 0.1, respectively, and train for \(R = 200\) communication rounds. To compute relevance-based attributions, we evaluate four attribution methods in our FL framework: Layerwise Relevance Propagation (FedXLRP) \cite{bach2015pixel}, Integrated Gradients (FedXIG) \cite{integrated_grad}, SmoothGrad (FedXSG) \cite{smoothgrad}, and Input\(\times\)Gradient (FedXGrad). Unless otherwiswe state we use  $\varepsilon=20$ sparsity level $s=70\%$, and $\lambda=0.5$. 
\subsection{Baselines}
The experimental results in \autoref{tab:results} shows that FedXLRP consistently outperforms all baselines across varying client numbers and data distributions. This advantage amplifies in more challenging scenarios (K=100, $\alpha$=0.05). FedFed ranks second, while FedAvg and FedAux degrade significantly with increased client numbers. On CIFAR-100, performance gaps narrow, though FedXLRP maintains its edge, particularly with higher client counts and heterogeneity. Other attribution methods (FedXIG, FedXGrad, FedXSG) perform well but remain below FedXLRP, highlighting attribution method choice importance. We discuss this point in Subsection \ref{sec:xai_compariosn}. 
In addtion, communication efficiency results \autoref{tab:comm_rounds} reinforce FedXLRP's advantages. To reach 70\% accuracy on CIFAR-10 with 10 clients, FedXLRP requires only 14 rounds versus 49 for FedAvg and 28 for FedFTG. With 100 clients, FedXLRP needs 23 rounds to reach 60\% accuracy compared to FedAvg's 79. CIFAR-100 shows similar patterns, with FedXLRP achieving target accuracies in substantially fewer rounds across client configurations. Among baselines, only FedFed shows comparable efficiency.
\begin{table*}[t]
\centering
\renewcommand{\arraystretch}{1.2}
\caption{Performance comparison across different federated learning methods, including standard deviations from 5 runs. Results show the top-1 test accuracy (\%) on CIFAR-10, CIFAR-100, and Tiny-ImageNet under varying numbers of clients (K) and Dirichlet concentration parameters ($\alpha$). Bold values indicate the best performance in each column.}
\label{tab:results}
\resizebox{\textwidth}{!}{%
\begin{tabular}{lcccccccccccc}
\toprule
Dataset & \multicolumn{4}{c}{CIFAR-10} & \multicolumn{4}{c}{CIFAR-100} & \multicolumn{4}{c}{Tiny-ImageNet} \\
\midrule
 & \multicolumn{2}{c}{K=10} & \multicolumn{2}{c}{K=100} & \multicolumn{2}{c}{K=10} & \multicolumn{2}{c}{K=100} & \multicolumn{2}{c}{K=10} & \multicolumn{2}{c}{K=100} \\
\midrule
& $\alpha=0.05$ & $\alpha=0.1$ & $\alpha=0.05$ & $\alpha=0.1$ & $\alpha=0.05$ & $\alpha=0.1$ & $\alpha=0.05$ & $\alpha=0.1$ & $\alpha=0.05$ & $\alpha=0.1$ & $\alpha=0.05$ & $\alpha=0.1$ \\
\midrule
FedAvg   & 64.16 $\pm$ 0.71 & 72.90 $\pm$ 0.55 & 55.53 $\pm$ 0.82 & 60.94 $\pm$ 0.68 & 43.82 $\pm$ 0.45 & 44.57 $\pm$ 0.51 & 32.99 $\pm$ 0.73 & 35.59 $\pm$ 0.62 & 30.30 $\pm$ 0.59 & 34.41 $\pm$ 0.48 & 26.20 $\pm$ 0.81 & 28.79 $\pm$ 0.75 \\
FedProx  & 71.56 $\pm$ 0.65 & 74.31 $\pm$ 0.49 & 51.89 $\pm$ 0.91 & 65.01 $\pm$ 0.53 & 44.33 $\pm$ 0.38 & 48.67 $\pm$ 0.47 & 37.79 $\pm$ 0.55 & 38.43 $\pm$ 0.49 & 33.27 $\pm$ 0.42 & 34.54 $\pm$ 0.51 & 24.15 $\pm$ 0.88 & 30.23 $\pm$ 0.67 \\
FedDyn   & 70.62 $\pm$ 0.58 & 78.25 $\pm$ 0.34 & 63.09 $\pm$ 0.61 & 69.41 $\pm$ 0.45 & 48.53 $\pm$ 0.41 & 49.42 $\pm$ 0.39 & 37.14 $\pm$ 0.68 & 41.60 $\pm$ 0.54 & 33.03 $\pm$ 0.48 & 35.69 $\pm$ 0.41 & 29.52 $\pm$ 0.63 & 32.37 $\pm$ 0.59 \\
SCAFFOLD & 71.35 $\pm$ 0.61 & 75.62 $\pm$ 0.42 & 58.51 $\pm$ 0.79 & 63.82 $\pm$ 0.62 & 46.61 $\pm$ 0.33 & 47.27 $\pm$ 0.44 & 35.28 $\pm$ 0.24 & 38.87 $\pm$ 0.58 & 31.53 $\pm$ 0.53 & 35.76 $\pm$ 0.39 & 28.35 $\pm$ 0.59 & 30.49 $\pm$ 0.61 \\
FedSAM   & 70.15 $\pm$ 0.88 & 77.53 $\pm$ 0.51 & 62.45 $\pm$ 0.72 & 68.78 $\pm$ 0.59 & 47.95 $\pm$ 0.35 & 48.81 $\pm$ 0.41 & 36.88 $\pm$ 0.51 & 40.92 $\pm$ 0.47 & 32.21 $\pm$ 0.55 & 35.03 $\pm$ 0.44 & 28.98 $\pm$ 0.66 & 31.84 $\pm$ 0.52 \\
FedDISCO & 69.73 $\pm$ 0.44 & 76.98 $\pm$ 0.63 & 61.82 $\pm$ 0.68 & 68.11 $\pm$ 0.65 & 47.31 $\pm$ 0.52 & 48.15 $\pm$ 0.58 & 36.17 $\pm$ 0.67 & 40.25 $\pm$ 0.61 & 31.78 $\pm$ 0.61 & 34.68 $\pm$ 0.52 & 28.51 $\pm$ 0.71 & 31.33 $\pm$ 0.64 \\
FedFed   & 79.12 $\pm$ 0.48 & 82.58 $\pm$ 0.31 & 75.35 $\pm$ 0.57 & 78.87 $\pm$ 0.43 & \textbf{52.27 $\pm$ 0.49} & 56.45 $\pm$ 0.38 & 45.41 $\pm$ 0.76 & 49.28 $\pm$ 0.52 & 34.99 $\pm$ 0.45 & 36.49 $\pm$ 0.37 & 33.28 $\pm$ 0.58 & \textbf{34.89 $\pm$ 0.47} \\
FedFTG   & 75.21 $\pm$ 0.53 & 78.44 $\pm$ 0.39 & 70.84 $\pm$ 0.31 & 74.85 $\pm$ 0.41 & 44.91 $\pm$ 0.62 & 54.15 $\pm$ 0.45 & 40.43 $\pm$ 0.39 & 46.47 $\pm$ 0.48 & 32.40 $\pm$ 0.57 & 33.79 $\pm$ 0.49 & 30.53 $\pm$ 0.51 & 32.26 $\pm$ 0.55 \\
FedGen   & 68.34 $\pm$ 0.68 & 74.65 $\pm$ 0.47 & 58.90 $\pm$ 0.75 & 63.20 $\pm$ 0.66 & 44.29 $\pm$ 0.55 & 50.12 $\pm$ 0.51 & 36.70 $\pm$ 0.62 & 39.82 $\pm$ 0.57 & 34.17 $\pm$ 0.44 & 36.33 $\pm$ 0.42 & 29.45 $\pm$ 0.68 & 31.60 $\pm$ 0.61 \\
FedAux   & 59.72 $\pm$ 0.85 & 73.70 $\pm$ 0.51 & 53.13 $\pm$ 0.88 & 65.11 $\pm$ 0.59 & 44.05 $\pm$ 0.48 & 44.85 $\pm$ 0.53 & 36.54 $\pm$ 0.65 & 38.39 $\pm$ 0.61 & 27.83 $\pm$ 0.72 & 34.31 $\pm$ 0.49 & 24.75 $\pm$ 0.82 & 30.38 $\pm$ 0.69 \\
FedDF    & 60.56 $\pm$ 0.77 & 72.40 $\pm$ 0.58 & 59.20 $\pm$ 0.72 & 63.47 $\pm$ 0.64 & 30.13 $\pm$ 0.81 & 45.47 $\pm$ 0.50 & 34.46 $\pm$ 0.69 & 36.77 $\pm$ 0.64 & 29.07 $\pm$ 0.65 & 34.75 $\pm$ 0.46 & 28.42 $\pm$ 0.62 & 30.51 $\pm$ 0.63 \\
\midrule
FedXLRP  & \textbf{81.72 $\pm$ 0.35} & \textbf{83.46 $\pm$ 0.28} & \textbf{77.02 $\pm$ 0.63} & \textbf{80.27 $\pm$ 0.39} & 52.07 $\pm$ 0.45 & \textbf{58.09 $\pm$ 0.33} & \textbf{46.25 $\pm$ 0.82} & \textbf{52.63 $\pm$ 0.41} & \textbf{36.85 $\pm$ 0.39} & \textbf{38.64 $\pm$ 0.32} & \textbf{34.64 $\pm$ 0.52} & 33.18 $\pm$ 0.51 \\
FedXIG   & 72.22 $\pm$ 0.59 & 75.89 $\pm$ 0.44 & 64.02 $\pm$ 0.69 & 68.61 $\pm$ 0.55 & 49.21 $\pm$ 0.48 & 55.46 $\pm$ 0.41 & 42.34 $\pm$ 0.58 & 47.89 $\pm$ 0.50 & 33.59 $\pm$ 0.51 & 35.30 $\pm$ 0.45 & 29.77 $\pm$ 0.64 & 31.93 $\pm$ 0.58 \\
FedXGrad & 71.38 $\pm$ 0.62 & 73.99 $\pm$ 0.52 & 64.52 $\pm$ 0.65 & 66.49 $\pm$ 0.61 & 50.13 $\pm$ 0.43 & 54.21 $\pm$ 0.46 & 42.11 $\pm$ 0.60 & 46.98 $\pm$ 0.53 & 33.18 $\pm$ 0.54 & 34.40 $\pm$ 0.48 & 30.00 $\pm$ 0.61 & 30.93 $\pm$ 0.60 \\
FedXSG   & 71.89 $\pm$ 0.60 & 74.63 $\pm$ 0.49 & 63.75 $\pm$ 0.71 & 67.80 $\pm$ 0.58 & 51.57 $\pm$ 0.39 & 55.22 $\pm$ 0.42 & 40.91 $\pm$ 0.63 & 47.61 $\pm$ 0.51 & 33.43 $\pm$ 0.52 & 34.74 $\pm$ 0.47 & 29.64 $\pm$ 0.65 & 31.53 $\pm$ 0.59 \\
\bottomrule
\end{tabular}%
}
\end{table*}
\begin{table*}[t]
\centering
\renewcommand{\arraystretch}{1.2}
\caption{Communication efficiency comparison across different federated learning methods. Results show the number of communication rounds (mean ± std over 5 runs) needed to achieve target accuracy thresholds (70\% and 60\% for CIFAR-10; 40\% and 30\% for CIFAR-100; 35\% and 30\% for Tiny-ImageNet) with different numbers of clients (K=10 and K=100) and heterogeneity parameter $\alpha=0.1$. \textbf{Lower} values indicate better communication efficiency. Best results are in bold.}
\label{tab:comm_rounds}
\footnotesize 
\begin{tabular}{lcccccc}
\toprule
 & \multicolumn{2}{c}{CIFAR-10} & \multicolumn{2}{c}{CIFAR-100} & \multicolumn{2}{c}{Tiny-ImageNet} \\
\cmidrule(lr){2-3} \cmidrule(lr){4-5} \cmidrule(lr){6-7}
 & \textit{acc}=70\% & \textit{acc}=60\% & \textit{acc}=40\% & \textit{acc}=30\% & \textit{acc}=35\% & \textit{acc}=30\% \\
Method & K=10 & K=100 & K=10 & K=100 & K=10 & K=100 \\
\midrule
FedAvg   & 49 $\pm$ 5 & 79 $\pm$ 7 & 24 $\pm$ 4 & 34 $\pm$ 5 & 60 $\pm$ 6 & 90 $\pm$ 8 \\
FedProx  & 32 $\pm$ 3 & 55 $\pm$ 4 & 28 $\pm$ 4 & 32 $\pm$ 4 & 45 $\pm$ 4 & 65 $\pm$ 5 \\
FedDyn   & 29 $\pm$ 3 & 51 $\pm$ 4 & 24 $\pm$ 3 & 32 $\pm$ 3 & 48 $\pm$ 5 & 60 $\pm$ 4 \\
SCAFFOLD & 33 $\pm$ 4 & 52 $\pm$ 5 & 25 $\pm$ 3 & 38 $\pm$ 4 & 46 $\pm$ 4 & 69 $\pm$ 6 \\
FedSAM   & 34 $\pm$ 4 & 58 $\pm$ 5 & 27 $\pm$ 4 & 35 $\pm$ 4 & 50 $\pm$ 5 & 64 $\pm$ 5 \\
FedDISCO & 36 $\pm$ 3 & 61 $\pm$ 6 & 28 $\pm$ 4 & 37 $\pm$ 5 & 52 $\pm$ 5 & 67 $\pm$ 6 \\
FedFed   & 15 $\pm$ 2 & \textbf{22 $\pm$ 4} & 13 $\pm$ 4 & \textbf{12 $\pm$ 3} & 12 $\pm$ 2 & 15 $\pm$ 2 \\
FedFTG   & 28 $\pm$ 3 & 57 $\pm$ 4 & 22 $\pm$ 3 & 28 $\pm$ 3 & 30 $\pm$ 3 & 40 $\pm$ 4 \\
FedGen   & 51 $\pm$ 5 & 66 $\pm$ 6 & 21 $\pm$ 4 & 39 $\pm$ 5 & 55 $\pm$ 5 & 70 $\pm$ 6 \\
FedAux   & 38 $\pm$ 4 & 45 $\pm$ 4 & 19 $\pm$ 3 & 36 $\pm$ 4 & 42 $\pm$ 4 & 50 $\pm$ 5 \\
FedDF    & 84 $\pm$ 7 & 90 $\pm$ 8 & 28 $\pm$ 4 & 45 $\pm$ 5 & 100 $\pm$ 8 & 110 $\pm$ 9 \\
\midrule
FedXLRP  & \textbf{14 $\pm$ 2} & 23 $\pm$ 3 & \textbf{11 $\pm$ 3} & 13 $\pm$ 2 & \textbf{10 $\pm$ 4} & \textbf{12 $\pm$ 2} \\
FedXIG   & 35 $\pm$ 4 & 39 $\pm$ 4 & 16 $\pm$ 2 & 21 $\pm$ 3 & 38 $\pm$ 4 & 45 $\pm$ 4 \\
FedXGrad & 41 $\pm$ 4 & 46 $\pm$ 5 & 13 $\pm$ 2 & 26 $\pm$ 3 & 44 $\pm$ 4 & 50 $\pm$ 5 \\
FedXSG   & 41 $\pm$ 5 & 51 $\pm$ 4 & 21 $\pm$ 3 & 33 $\pm$ 4 & 42 $\pm$ 4 & 48 $\pm$ 5 \\
\bottomrule
\end{tabular}%
\end{table*}

\subsection{Experiments on Real-World Datasets}
We also evaluate FedXDS on real-world datasets using implementations from the LEAF \cite{leaf} and \cite{pfl-bench} frameworks. Specifically, we consider CelebA and FEMNIST, which exhibit a distinct form of non-IIDness known as distribution shift, introducing an additional challenge for federated learning. For CelebA, we follow the implementation of \cite{FedGen}, while for FEMNIST, we use the setup from \cite{pfl-bench}, conducting experiments with 10 and 100 clients, respectively. Additional implementation details can be found in the supplementary material.
\begin{table}[t]
\centering
\caption{Accuracy comparison (mean ± std \% over 5 runs) on CelebA and FEMNIST. Best results are in bold. We use the LRP variant of FedXDS.}
\small 
\begin{tabular}{lcc}
\toprule
Method    & CelebA  & FEMNIST  \\
\midrule
FedAvg    & 87.23 $\pm$ 0.71 & 84.71 $\pm$ 0.68 \\
FedProx   & 87.79 $\pm$ 0.65 & 85.66 $\pm$ 0.61 \\
SCAFFOLD  & 86.36 $\pm$ 0.56 & 84.24 $\pm$ 0.59 \\
FedDyn    & 88.13 $\pm$ 0.52 & 86.45 $\pm$ 0.34 \\
FedSAM    & 90.03 $\pm$ 0.15 & 87.56 $\pm$ 0.44 \\
FedDISCO  & 89.78 $\pm$ 0.61 & 87.29 $\pm$ 0.62 \\
FedFTG    & 90.31 $\pm$ 0.43 & 87.88 $\pm$ 0.67 \\
FedGen    & 89.65 $\pm$ 0.59 & 86.37 $\pm$ 0.55 \\
FedDF     & 88.56 $\pm$ 0.63 & 85.28 $\pm$ 0.66 \\
FedAux    & 89.27 $\pm$ 0.60 & 85.94 $\pm$ 0.62 \\
FedFed    & 90.76 $\pm$ 0.83 & 88.28 $\pm$ 0.71 \\
\midrule
FedXDS    & \textbf{91.55 $\pm$ 0.48} & \textbf{89.03 $\pm$ 0.35} \\
\bottomrule
\end{tabular}
\label{tab:celeba_femnist}
\end{table}
The results in Table~\ref{tab:celeba_femnist} show a clear trend where FedXDS consistently achieves the highest accuracy across both datasets, demonstrating its effectiveness in handling distribution shift. Notably, methods incorporating additional data generation or augmentation, such as FedFTG and FedGen, also perform well, but FedXDS surpasses them, suggesting that relevance-guided data sharing provides a stronger mechanism for improving generalization in heterogeneous federated settings.


\begin{table}[t]
\centering
\caption{Performance comparison (mean ± std \% over 5 runs) on CIFAR-10 for different base FL methods and their FedXDS (using LRP) variants under two heterogeneity settings ($K=10$ and $K=100$). The FedXDS variants consistently improve performance.}
\small 
\begin{tabular}{l|cc}
\toprule
\textbf{Method} & $K=10$  & $ K=100$ \\ \midrule
FedAvg                 & 72.90 $\pm$ 0.55  & 60.94 $\pm$ 0.68 \\
\textbf{FedAvg + FedXDS} & \textbf{83.46 $\pm$ 0.31}  & \textbf{80.29 $\pm$ 0.42} \\ \midrule
FedProx                & 74.31 $\pm$ 0.49  & 65.01 $\pm$ 0.53 \\
\textbf{FedProx + FedXDS} & \textbf{84.05 $\pm$ 0.35}  & \textbf{80.73 $\pm$ 0.45} \\ \midrule
FedDyn                 & 70.62 $\pm$ 0.58  & 69.41 $\pm$ 0.45 \\
\textbf{FedDyn + FedXDS} & \textbf{84.20 $\pm$ 0.33}  & \textbf{79.56 $\pm$ 0.48} \\ \midrule
SCAFFOLD               & 71.35 $\pm$ 0.61  & 63.82 $\pm$ 0.62 \\
\textbf{SCAFFOLD + FedXDS} & \textbf{83.89 $\pm$ 0.38}  & \textbf{79.72 $\pm$ 0.44} \\ \bottomrule
\end{tabular}
\label{tab:fedxds+other}
\end{table}
The results in \autoref{tab:fedxds+other} demonstrate that integrating FedXDS with FL methods consistently improves the performance. In all cases, the addition of FedXDS yields substantial accuracy gains, particularly in the highly heterogeneous setting with K=100 clients. For instance, FedAvg, which traditionally suffers from performance degradation in non-IID scenarios, improves from 60.94\% to 80.29\% when combined with FedXDS. Similar improvements are observed for FedProx, FedDyn, and SCAFFOLD, with FedXDS enhancing their robustness to data heterogeneity while maintaining their inherent advantages.

The magnitude of improvement suggests that FedXDS effectively mitigates the adverse effects of statistical heterogeneity by facilitating better knowledge transfer across clients. Notably, FedDyn, which incorporates regularization-based local adaptation, exhibits the smallest absolute improvement with FedXDS, indicating that its existing adaptation mechanisms already partially address heterogeneity. In contrast, methods like FedAvg, without correction for client drift, benefit more from FedXDS.
\section{Empircal Privacy Analysis}
\label{sec:privacy_analysis}
\begin{figure*}[ht!]
    \centering
    \begin{subfigure}[t]{0.24\textwidth}
        \centering
        \includegraphics[width=\linewidth]{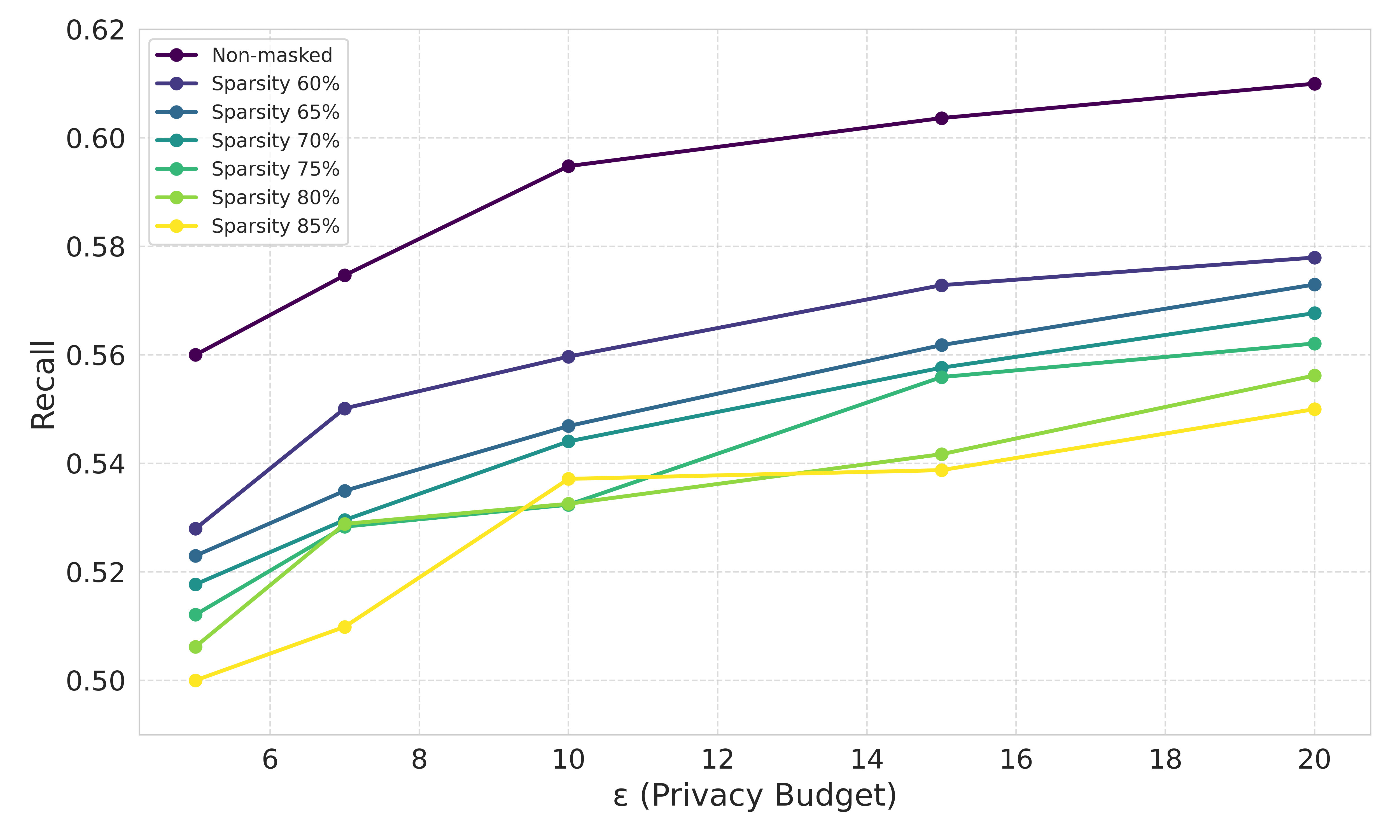}
        \caption{Recall score of MI attack at different $\varepsilon$ for various sparsity levels $s$.}
        \label{fig:MI_attack}
    \end{subfigure}
    \hfill
    \begin{subfigure}[t]{0.24\textwidth}
        \centering
        \includegraphics[width=\linewidth]{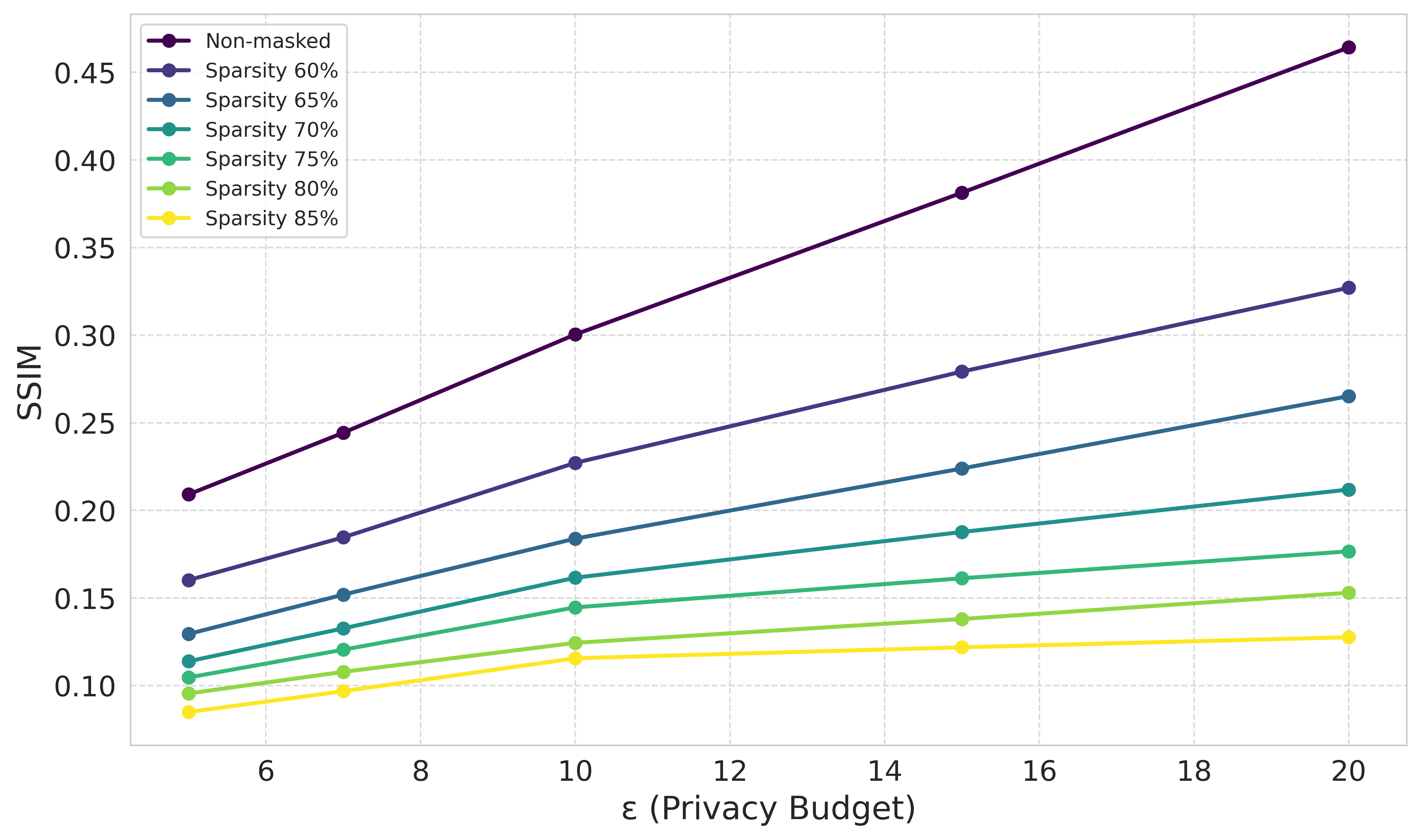}
        \caption{SSIM score of denoising attack at different $\varepsilon$ and sparsity levels $s$.}
        \label{fig:ssim_inversion}
    \end{subfigure}
    \hfill
    \begin{subfigure}[t]{0.24\textwidth}
        \centering
        \includegraphics[width=\linewidth]{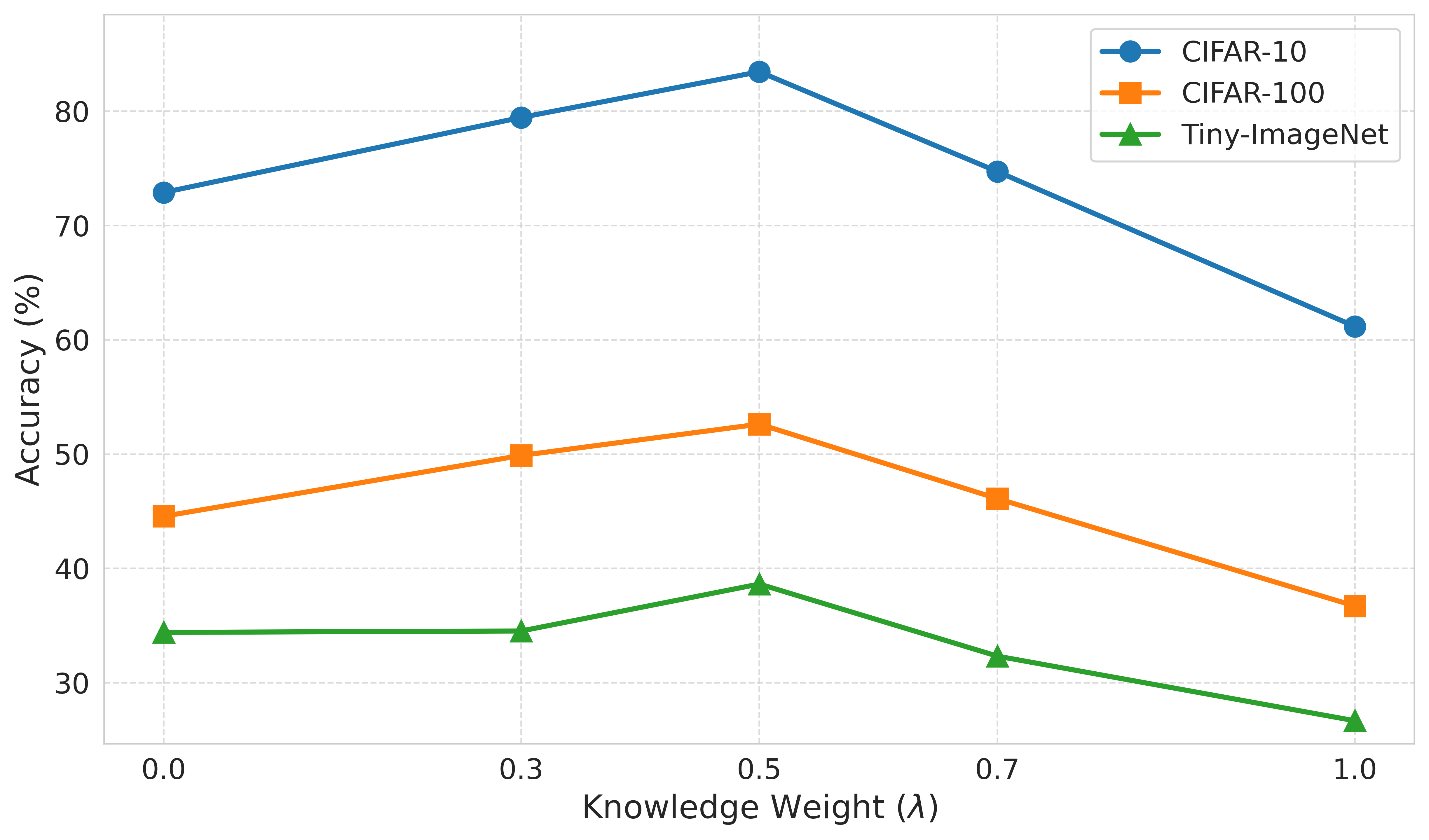}
        \caption{Effect of knowledge weight $\lambda$ on model accuracy across datasets.}
        \label{fig:lambda_ablation}
    \end{subfigure}
    \hfill
    \begin{subfigure}[t]{0.24\textwidth}
        \centering
        \includegraphics[width=\linewidth]{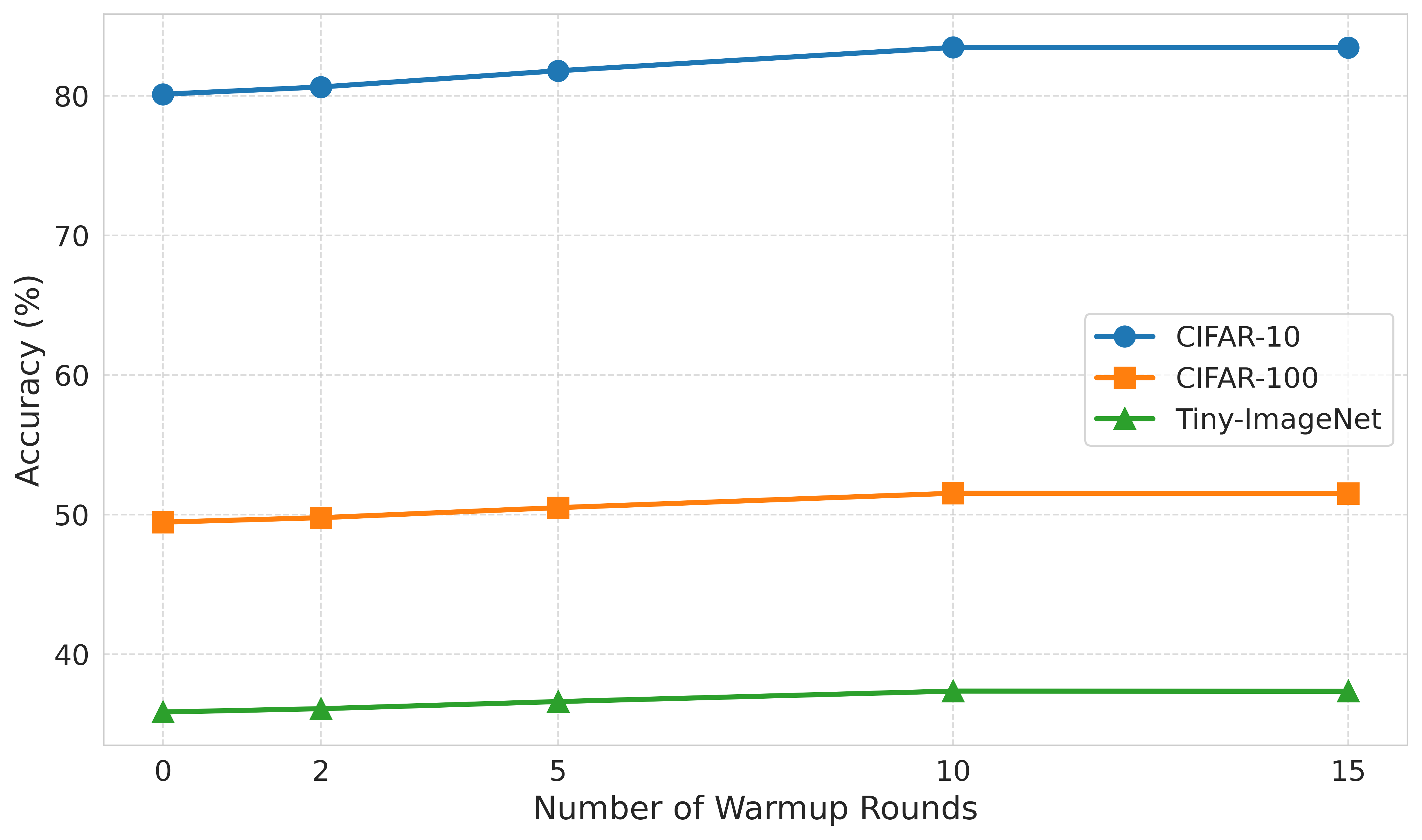}
        \caption{Impact of warmup rounds on model accuracy across datasets.}
        \label{fig:warmup_rounds}
    \end{subfigure}
    \caption{Evaluation of privacy-utility trade-offs and hyperparameter effects in our framework.}
    \label{fig:combined}
\end{figure*}

\subsection{Membership Inference Attack}
We conduct membership inference attacks following \cite{FedFed} to evaluate our methodology's privacy guarantees. We investigate whether training on masked and noised features reveals membership in the underlying dataset by using the globally DP data to train a shadow model, then testing if clients can infer membership by querying this model and training a random forest classifier. \autoref{fig:MI_attack} shows recall values for different privacy parameters $\varepsilon$. Results demonstrate that non-masked features, despite DP protection, offer inferior privacy protection compared to masked data. This can be explained by the fact that sparsity significantly mitigates attacks through norm reduction from attribution masks (\autoref{eq:binary_mask}). 
\subsection{Feature Inversion Attack}
While metric private features prevent direct inference of individual data features, a sophisticated adversary might train a denoising autoencoder to learn mappings between protected representations and original images. We conduct feature inversion attacks and report the SSIM  score between original and reconstructed images. \autoref{fig:ssim_inversion} reveals inversion attack success increases with lower sparsity and higher privacy budgets. Unmasked features consistently exhibit the highest SSIM, confirming that raw features under identical DP constraints provide inferior protection compared to our approach. Additional validations in supplementary material, including \textit{visualization} of inversion results for non-masked features and FedFed \cite{FedFed} features, confirming both cases' vulnerability to reconstruction, whereas our method provides superior protection. 
\section{Ablation Studies and Discussion}

\subsection{XAI Method Comparison Across Sparsification}
\label{sec:xai_compariosn}
\begin{table}[t]
\centering
\renewcommand{\arraystretch}{1.0}
\caption{Test accuracy (mean ± std \% over 5 runs) for different sparsification levels $s$ across various attribution methods. Best results are in bold. Results for CIFAR10 at $K=10, \alpha=0.1$.}
\label{tab:acc_vs_sparse}
\resizebox{\columnwidth}{!}{%
\begin{tabular}{rcccc}
\toprule
$s$ & Gradient & SmoothGrad & Int. Grad. & LRP \\
\midrule
60 & 75.99 $\pm$ 0.52 & 75.63 $\pm$ 0.55 & 77.10 $\pm$ 0.48 & \textbf{84.74 $\pm$ 0.31} \\
65 & 74.19 $\pm$ 0.56 & 73.15 $\pm$ 0.61 & 76.31 $\pm$ 0.51 & \textbf{84.01 $\pm$ 0.35} \\
70 & 73.99 $\pm$ 0.59 & 74.63 $\pm$ 0.54 & 75.89 $\pm$ 0.49 & \textbf{83.46 $\pm$ 0.28} \\
75 & 70.54 $\pm$ 0.65 & 70.67 $\pm$ 0.63 & 71.68 $\pm$ 0.58 & \textbf{83.01 $\pm$ 0.33} \\
80 & 68.31 $\pm$ 0.71 & 70.45 $\pm$ 0.66 & 70.98 $\pm$ 0.62 & \textbf{82.67 $\pm$ 0.36} \\
85 & 68.80 $\pm$ 0.68 & 68.13 $\pm$ 0.73 & 69.73 $\pm$ 0.67 & \textbf{81.88 $\pm$ 0.41} \\
\bottomrule
\end{tabular}%
}
\end{table}
Table~\ref{tab:acc_vs_sparse} demonstrates that FedXDS maintains stable performance across different sparsification levels, with all attribution methods achieving above 68\% accuracy, even at high sparsity thresholds. Notably, LRP consistently outperforms other methods, preserving accuracies between 81.88\% and 84.74\%, while others degrade more sharply under extreme sparsification. LRP's robustness stems from its layer-wise relevance propagation mechanism, which is fundamentally different from gradient-based approaches. Instead of measuring local sensitivity, LRP redistributes the model's output prediction backward through the network, layer by layer, conserving relevance at each step. This process aggregates relevance into contiguous, semantically coherent regions that represent structured object parts. In stark contrast, methods like SmoothGrad or Integrated Gradients identify 'hotspot' pixels that are locally influential but often disconnected from each other. Consequently, under high sparsification, LRP retains meaningful visual patterns, whereas other methods are left with a sparse collection of isolated points that lack sufficient context for accurate classification. We conclude that \emph{structural coherence} in saliency maps is an essential criterion when selecting the attribution method. This finding is corroborated by current literature \cite{LRP-Review, LPR-coherence, xai-benchmark}, which confirms LRP's superior explanation quality and reinforces that our work is consistent with established findings on the importance of explanation quality for downstream tasks. More discussion can be found in the supplementary material.
\subsection{Privacy vs. Utility}
We investigate how the privacy budget $\varepsilon$ impacts model performance across attribution methods. \autoref{tab:privacy_utility} shows that increasing $\varepsilon$ improves accuracy for all methods, demonstrating the expected privacy-utility trade-off: stronger privacy (lower $\varepsilon$) leads to reduced performance due to greater noise. Among all methods, LRP consistently achieves the highest accuracy across privacy levels outperforming the next best (Integrated Gradients) by 7-8 \%  and maintaining robustness even under strict privacy constraints. SmoothGrad shows modest improvements over basic gradients, though less pronounced than Integrated Gradients. These results underscore LRP's ability to extract stable, task-relevant features leading to better generalization with less privacy leakage, aligning with prior work on explanation fidelity~\cite{vis_lrp_faithful, bach2015pixel}.

\begin{table}[t]
\centering
\renewcommand{\arraystretch}{1.0}
\caption{Ablation study showing test accuracy ) for different privacy budgets ($\varepsilon$) across XAI-based federated learning methods. Results for CIFAR10 at $K=10, \alpha=0.1$.}
\label{tab:privacy_utility}
\resizebox{\columnwidth}{!}{%
\begin{tabular}{rcccc}
\toprule
$\varepsilon$ & Gradient & SmoothGrad & Int. Grad. & LRP \\
\midrule
5  & 70.55 $\pm$ 0.62 & 71.23 $\pm$ 0.59 & 72.45 $\pm$ 0.52 & \textbf{80.06 $\pm$ 0.33} \\
7  & 71.45 $\pm$ 0.58 & 72.15 $\pm$ 0.55 & 73.35 $\pm$ 0.47 & \textbf{80.89 $\pm$ 0.31} \\
10 & 72.25 $\pm$ 0.51 & 72.95 $\pm$ 0.48 & 74.15 $\pm$ 0.41 & \textbf{81.65 $\pm$ 0.29} \\
15 & 73.12 $\pm$ 0.45 & 73.85 $\pm$ 0.43 & 75.02 $\pm$ 0.38 & \textbf{82.11 $\pm$ 0.25} \\
20 & 73.99 $\pm$ 0.42 & 74.63 $\pm$ 0.39 & 75.89 $\pm$ 0.35 & \textbf{83.46 $\pm$ 0.28} \\
\bottomrule
\end{tabular}%
}
\end{table}


\subsection{Ablation Study on the Knowledge Weight}
We evaluate how the knowledge weight $\lambda$, as defined in \autoref{eq:fedxds_learning_objective}, affects performance across CIFAR-10, CIFAR-100, and Tiny-ImageNet. \autoref{fig:lambda_ablation} shows that intermediate values ($\lambda=0.5$) consistently outperform both extremes, with CIFAR-10 accuracy peaking at 83.46\% compared to 72.90\% ($\lambda=0.0$, equivalent to FedAvg) and 61.18\% ($\lambda=1.0$, synthetic data equal weight). Our ablation thus shows that balancing the contriubution of synthetic data is critical for performance.
\subsection{Warmup and Efficiency}
\autoref{fig:warmup_rounds} shows accuracy slightly improves with more warmup rounds, plateauing at 10, suggesting this is the optimal trade-off between convergence and communication cost. Our approach is also computationally efficient. Attribution masks are computed in a single backward pass during warmup, incurring minimal overhead. In contrast, methods like FedGen and FedFTG retrain generators every round, and FedFed requires costly VAE training—making FedXDS more suitable for resource-constrained settings. We provide more discussion in the supplementary material.
\section{Conclusion}
In this work, we introduced FedXDS, leveraging attribution methods to address statistical heterogeneity, privacy, and computational efficiency in federated learning. By using attribution maps to extract and share privacy-preserved task-relevant features, our approach enables effective knowledge transfer while ensuring strong privacy guarantees. Experiments show that FedXDS consistently outperforms existing methods in accuracy and communication efficiency. Moreover, the sparsity induced by relevance attribution enhances privacy, mitigating feature inversion and membership inference risks. These findings highlight attribution methods as a promising tool for improving federated learning while preserving privacy and efficiency.

\section{Acknowledgements}
This work was supported by the European Union’s Horizon Europe research and innovation programme (EU Horizon Europe) as grant ACHILLES (101189689).
{
    \small
    \bibliographystyle{ieeenat_fullname}
    \bibliography{main}
}

\clearpage
\appendix
\twocolumn[{%
\centering
\Large\textbf{Supplementary Material\\FedXDS: Leveraging Model Attribution Methods to Counteract Data Heterogeneity in Federated Learning}\\[1.5em]
}]

\section{Implementation Details}
All models use a ResNet8 architecture and are trained with stochastic gradient descent (SGD) with a momentum of 0.9 and a learning rate of 0.01. Attribution computations are performed using the Zennit library \cite{zennit}. For evaluation, we use top-1 accuracy as the primary metric.

\textbf{FEMNIST} For our implementation, we use the FEMNIST dataset as organized in Caldas et al. (2019). This dataset, derived from EMNIST, includes 62 classes (digits and both uppercase and lowercase letters). Each client’s data comes from a single writer—roughly 200–300 samples per client. We work with 100 clients in total and select 10 clients per communication round.

\textbf{CelebA} We evaluate our method on CelebA, following the setup of \cite{FedGen} and \cite{leaf}. The dataset is structured as a binary classification task, distinguishing between "smiling" and "not smiling." We utilize ten clients with a 50\% participation rate per round. This dataset is important from the vantage point of privacy. We show below reconstruction attacks on faces, trying to re-identify faces visually.

\section{Extended Discussion on Task-Relevant Features and Privacy}
\label{app:privacy_discussion}

In our work, we build upon the finding that sharing features derived from raw client data helps counteract data heterogeneity in federated learning. However, this approach inherently poses privacy risks. While differential privacy (DP) could in principle be applied directly to the raw client data, the high dimensionality of image data would necessitate substantial noise addition, significantly decreasing utility.
$\varepsilon$-metric privacy. This method allows us to maintain strong privacy guarantees while preserving the utility benefits of feature sharing. Unlike standard differential privacy, which defines neighboring datasets by a single-entry difference, we define privacy based on input similarity under a metric $d_X$, allowing a more natural treatment of image data where task-relevant and potentially sensitive features (e.g., facial attributes, backgrounds) are often inseparable.

We emphasize that our approach treats all retained pixels after masking as equally sensitive and applies our privacy mechanism  to every pixel that remains after masking. The attribution-based method identifies these crucial regions dynamically without predefined sensitive regions, ensuring adaptive privacy without manual supervision. Notably, we do \textbf{not} assume that the \textbf{retained} pixels are "less private" than those discarded; rather, our $\varepsilon$-metric mechanism uniformly bounds the disclosure risk of whichever features are deemed relevant.
By removing uninformative regions altogether and applying noise only to the most discriminative subset, we reduce the dimensionality of data requiring protection while limiting an adversary's ability to infer private details from the shared features. This approach provides better theoretical guarantees while maintaining utility. Our aim is to protect against membership inference attacks (MIA) and feature inversion when sharing these masked features, which we demonstrate empirically in . Below we detail the technical arguments that ensure metric privacy.

\subsection{Why Task-Relevant is equivalent with Potentially Sensitive Features}

Many real-world image tasks involve visual features that are both \emph{critical for classification} and \emph{potentially identifying} (e.g., facial landmarks, unique physical traits, or distinctive backgrounds). While it might be tempting to assume that discarding “sensitive regions” a priori would suffice, this often fails in practice if the model itself relies on exactly those regions for accurate predictions. Our method \emph{does not} rely on any manual definition of what constitutes “sensitive” or “background.” Instead, we let an \emph{attribution-based mask} discover the truly \emph{discriminative} subset of pixels. Because these retained pixels can naturally include highly private elements (for example, a key part of the face), we \emph{treat all retained pixels as equally sensitive} and apply our privacy mechanism to every pixel that survives the mask.

\subsection{Adaptive Attribution Masking Without Manual Annotations}

The binary mask $\mathbf{m}$ arises from an attribution method that identifies pixels most influential to the model’s output. This ensures an \emph{adaptive}, data-driven selection of features, mitigating the need for \emph{pre-labeled} sensitive regions (e.g., bounding boxes for faces). By defaulting to the viewpoint that any discriminative pixel is potentially private, our approach sidesteps the risk of \emph{missing} an unintuitive but privacy-revealing region. All masked-out (zeroed) pixels are completely removed from subsequent sharing, effectively \emph{reducing} the dimensionality of data exposed to the privacy mechanism.

\subsection{Uniform Privacy Guarantee Over Retained Pixels}

One might worry that by selectively \emph{retaining} the most discriminative features, the mask could inadvertently highlight the \emph{most private} parts of the image (such as a person’s unique facial contour). However, under our $\varepsilon$-metric privacy model, the mechanism adds \emph{Gaussian noise} calibrated to a \emph{worst-case sensitivity} $\Delta_f = 1$. Concretely, for any two images that differ in potentially identifying ways, the \emph{distribution} of the noisy output remains within $e^{\varepsilon}$ of each other, bounding an adversary’s ability to infer whether a specific individual’s pixels were present. Hence, even if a user’s exact facial features remain in the retained region, the probability of distinguishing their face from another’s in the shared representation remains formally constrained by $\varepsilon$.

\subsection{Practical Advantages of Non-Expansive Masking}

As detailed in Section~4.2, our function $f(\mathbf{x}) = \mathbf{x} \odot \mathbf{m}$ is \emph{non-expansive} under the $\ell_2$ norm, yielding a bounded global sensitivity $\Delta_f \le 1$. This bounded sensitivity is crucial: it means the \emph{noise scale} we add does not have to grow arbitrarily for large or varied image spaces. By retaining only the task-relevant pixels and zeroing out the rest, we effectively:
\begin{itemize}
    \item Remove vast amounts of non-essential background or other context that might be re-identifying in unexpected ways.
    \item Limit how much noise is needed to preserve privacy, thereby improving the \emph{utility–privacy tradeoff}.
\end{itemize}

\subsection{Extensions and Future Directions}

\paragraph{Region-Specific Noise}
While we apply a \emph{single} noise scale $\sigma$ to all retained pixels, the same framework could be extended to \emph{varying} noise scales for different groups of pixels if domain knowledge indicates that some regions (e.g., near eyes) are more sensitive. This would require refining the sensitivity analysis (e.g., weighting the mask) but is straightforwardly encompassed by the metric privacy formulation.

\paragraph{Combining with Other Privacy Filters}
In high-stakes settings where certain features (like license plates or medical indicators) are strictly off-limits, one could combine our attribution mask with \emph{rule-based filters} that forcibly zero out specific known identifiers before applying noise. This hybrid approach ensures that critical known-sensitive areas are always removed, while other discriminative regions receive the DP-based protection automatically.

\paragraph{Adapting the Metric \(d_X\)}
We adopt a straightforward $\ell_2$ norm over image space for simplicity, but the metric $d_X$ can be customized (e.g., $\ell_1$, embedding-based distances, or perceptual metrics). The choice of metric can further align with specific privacy or robustness goals, such as controlling visually perceptible differences.


\section{Sensitivity for Differential Privacy}

\subsection{Mathematical Formulation}

In traditional differential privacy, the sensitivity of a function $f: \mathcal{D} \rightarrow \mathbb{R}^m$ measures how much the output can change when one record in the dataset changes. For continuous data domains like images, a more appropriate notion is the following senstivity choice as used in prior works \cite{test_lipschitz, Lipschitz-Definition, CI-lipschitz} as follows:

\begin{definition}[Sensitivity]
For a function $f: X \rightarrow \mathbb{R}^m$ where $(X, d_X)$ is a metric space, the Lipschitz sensitivity of $f$ is defined as:
\[
\Delta_f = \sup_{x, x' \in X, x \neq x'} \frac{\|f(x) - f(x')\|_2}{d_X(x, x')}
\]
\end{definition}

This definition captures the maximum rate of change of the function $f$ with respect to changes in the input space. A function with this sensitivity $\Delta_f$ satisfies:
\[
\|f(x) - f(x')\|_2 \leq \Delta_f \cdot d_X(x, x')
\]
for all $x, x' \in X$. In the context of image data, $X = \mathbb{R}^{H \times W \times C}$ typically represents the space of images with height $H$, width $W$, and $C$ channels, and $d_X(x, x') = \|x - x'\|_2$ is the Euclidean distance between images.

\subsection{Privacy Mechanism}

Using this sensitivity definition, we can construct a privacy mechanism by adding calibrated Gaussian noise:

\begin{equation}
\mathcal{M}(x) = f(x) + \mathcal{N}(0, \sigma^2 I_m)
\end{equation}

where $\sigma = \frac{\Delta_f \cdot \sqrt{2\ln(1.25/\delta)}}{\varepsilon}$ for $(\varepsilon, \delta)$-differential privacy. This mechanism satisfies:

\begin{theorem}
For a function $f: X \rightarrow \mathbb{R}^m$ with Lipschitz sensitivity $\Delta_f$, the mechanism $\mathcal{M}(x) = f(x) + \mathcal{N}(0, \sigma^2 I_m)$ with $\sigma$ as defined above is $(\varepsilon, \delta)$-differentially private with respect to the metric $d_X$.
\end{theorem}

\subsection{Practical Advantages of our Sensitivity Choice}

Our choise of sensitivity provides a natural way to reason about privacy for continuous domains like images, where small changes in input should correspond to proportionally small privacy losses, unlike traditional sensitivity which is designed for discrete changes. This approach exhibits scale invariance, automatically adjusting to the scale of the data—if inputs are scaled by some factor, the privacy guarantees remain consistent without needing to recalibrate the privacy mechanism.

A significant benefit is tighter noise calibration. For many applications, Lipschitz sensitivity allows adding noise proportional to the actual "size" of the input difference rather than adding a fixed amount of noise based on worst-case scenarios, leading to better utility. In image data, where the distance between inputs has semantic meaning (similar images are close in pixel space), this choice of sensitivity respects this semantic structure in the privacy guarantee.

For unbounded domains like $\mathbb{R}^d$, traditional global sensitivity might be infinite, rendering standard mechanisms unusable, while the sensitivity can remain finite. This approach also aligns well with stable machine learning algorithms that are already designed so that small input changes cause small output changes, making them naturally compatible with our sensitivity frameworks.

\subsection{Graduated Protection Based on Image Similarity}

Unlike binary notions of privacy, Lipschitz sensitivity provides a continuous spectrum of protection that aligns with visual similarity. This is formalized through the concept of indistinguishability:

\begin{proposition}[Graduated Protection]
For two images $x, x'$ with distance $d_X(x, x') = d$, and a mechanism $\mathcal{M}$ with Lipschitz sensitivity $\Delta_f$, the privacy loss is bounded by $\varepsilon \cdot d$.
\end{proposition}

This means that visually similar images (small $d$) are more difficult to distinguish even with access to the private output, while images that are perceptually very different receive appropriately scaled protection.

\subsubsection{Natural Handling of Feature Correlations}

Images exhibit strong spatial correlations between pixels, which traditional independent noise addition fails to respect. Lipschitz sensitivity naturally accommodates these correlations:

\begin{proposition}[Correlation Preservation]
For spatially correlated features in images, Lipschitz-based mechanisms preserve the relative importance of correlated structures while providing privacy guarantees.
\end{proposition}

This property ensures that important visual structures (edges, textures, shapes) receive appropriate protection without being disproportionately distorted by the privacy mechanism.

\subsection{Applications to Federated Learning}

In the context of federated learning with image data, Lipschitz sensitivity provides a theoretical foundation for:

\begin{enumerate}
\item Bounding information leakage from shared features
\item Ensuring that small variations in sensitive attributes cannot be recovered
\item Providing guarantees against membership inference attacks
\item Enabling privacy-utility tradeoffs that scale with the semantic importance of features
\end{enumerate}

The use of attribution-based masking in combination with Lipschitz sensitivity, as employed in our method, further enhances these properties by focusing the privacy protection on the most task-relevant features, ensuring that noise addition is maximally efficient in preserving utility while maintaining privacy guarantees.

\subsection{Comparison with Traditional Sensitivity}

Traditional global sensitivity for a function $f: \mathcal{D} \rightarrow \mathbb{R}^m$ is defined as:
\[
\Delta_f^{global} = \max_{D, D' \text{ adjacent}} \|f(D) - f(D')\|_2
\]

For unbounded continuous domains like images, this sensitivity can be infinite, making standard differential privacy mechanisms unusable. In contrast, Lipschitz sensitivity remains bounded as long as the function $f$ is Lipschitz continuous, which is the case for many practical feature extraction methods in computer vision.

Additionally, for high-dimensional data, traditional additive noise mechanisms require noise scaling with $\sqrt{d}$ for $d$-dimensional outputs, whereas Lipschitz sensitivity allows for noise calibration that depends only on the privacy parameters and the Lipschitz constant, not the dimensionality of the data.

\subsection{Sensitivity Analysis and Dimensionality Reduction}
\label{subsec:sensitivity_analysis}

An interesting observation regarding our sensitivity bound $\Delta_f \leq 1$ is that this bound equals the sensitivity of the identity function. Indeed, for the identity function $\text{id}(x) = x$, we have $\Delta_{\text{id}} = \max_{x,x'} \frac{\|x-x'\|}{\|x-x'\|} = 1$. This raises the question: If both our masking approach and the unmasked (identity) approach have the same formal sensitivity, what concrete privacy advantage does our method provide?

The key insight lies in the \emph{dimensionality} of the space requiring noise addition. Consider an image $\mathbf{x} \in \mathbb{R}^{H \times W \times 3}$ with $d = H \times W \times 3$ total dimensions. With the standard approach (no masking), noise would be added to all $d$ dimensions according to $\mathcal{M}_{\text{id}}(\mathbf{x}) = \mathbf{x} + \mathcal{N}(0, \sigma^2\mathbf{I})$. In contrast, our approach first applies a sparsifying mask $\mathbf{m}$ that retains only a fraction $\alpha = \frac{\|\mathbf{m}\|_0}{d}$ of the dimensions, where $\|\mathbf{m}\|_0$ counts the number of non-zero elements in $\mathbf{m}$.

For a given privacy budget $\varepsilon$, both approaches require the same per-dimension noise scale $\sigma$. However, the total expected squared $\ell_2$ distortion differs significantly:
\begin{align}
\mathbb{E}[\|\mathcal{M}_{\text{id}}(\mathbf{x}) - \mathbf{x}\|_2^2] &= d\sigma^2 \\
\mathbb{E}[\|\mathcal{M}(\mathbf{x}) - \mathbf{x}\|_2^2] &= \alpha d\sigma^2 + \|(\mathbf{1} - \mathbf{m}) \odot \mathbf{x}\|_2^2
\end{align}

The first term in our approach's distortion ($\alpha d\sigma^2$) represents the noise added to the retained dimensions, while the second term represents the distortion from setting masked dimensions to zero. Crucially, the masked dimensions are specifically those deemed less relevant to the task by our attribution method. This means that for a fixed privacy budget $\varepsilon$, our approach concentrates the noise budget on fewer, more task-relevant dimensions rather than spreading it across all dimensions.

Furthermore, by applying our attribution-guided masking, we explicitly remove potentially identifying background features entirely, rather than simply perturbing them with noise. This dual effect—reducing dimensionality while focusing on task-relevant features—allows our approach to maintain higher utility at equivalent privacy levels, as demonstrated empirically in the main text.

This analysis highlights a fundamental principle in privacy-preserving machine learning: when limited to a fixed privacy budget, selective disclosure of the most task-relevant features often yields better utility than indiscriminate disclosure of all features with uniformly distributed noise. Our attribution-guided masking approach provides a principled method for implementing this selective disclosure while maintaining formal privacy guarantees.
\section{Inversion Attacks}
\begin{figure*}[t]
    \centering
    \includegraphics[width=0.8\textwidth]{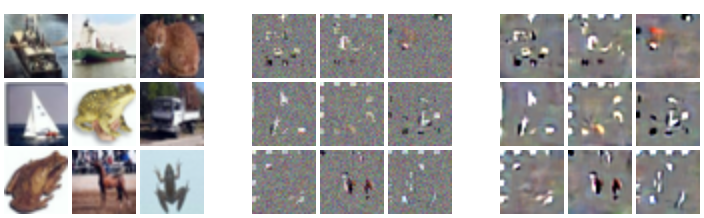}
    \caption{Inversion Attack on FedXDS}
    \label{fig:fedxds}
\end{figure*}

\begin{figure*}[t]
    \centering
    \includegraphics[width=0.8\textwidth]{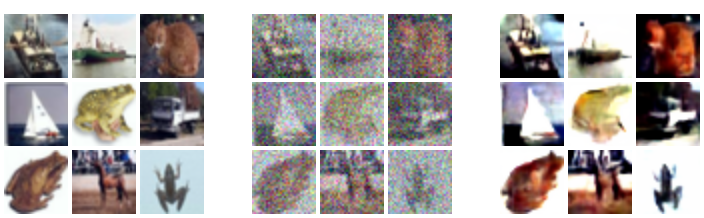}
    \caption{Inversion Attack on Raw Features}
    \label{fig:raw}
\end{figure*}

\begin{figure*}[t]
    \centering
    \includegraphics[width=0.8\textwidth]{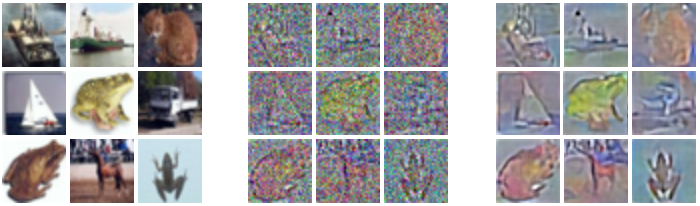}
    \caption{Inversion Attack on FedFed}
    \label{fig:fedfed}
\end{figure*}

We conduct inversion attacks on three different methods. Specifically we show the denoising inversion attack for our method FedXDS \autoref{fig:fedxds}, for the raw and DP protected features \autoref{fig:raw}, and for FedFed in \autoref{fig:fedfed} which shares DP distilled features \cite{FedFed}. The results shows the superiority of our method in finding the most crucial representations and protecting crucial features needed for generalization. In contrast, raw features heavily leak privacy under the same DP constraints. This speaks to the validity of our privacy argument, i.e., that the norm is significantly reduced such that we get a stronger DP guarantee at the same noise level. In addition, FedFed also appears to be prone to the denoising attack. Although not as severe as in sharing raw features, it is nevertheless possible to reconstruct a considerable amount from the original images. 

\begin{figure*}[t]
\centering
\begin{subfigure}[b]{0.22\textwidth}
    \centering
    \includegraphics[width=\textwidth]{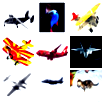}
    \caption{Original Images}
    \label{fig:original}
\end{subfigure}
\hfill
\begin{subfigure}[b]{0.22\textwidth}
    \centering
    \includegraphics[width=\textwidth]{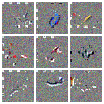}
    \caption{Protected Features}
    \label{fig:protected}
\end{subfigure}
\hfill
\begin{subfigure}[b]{0.22\textwidth}
    \centering
    \includegraphics[width=\textwidth]{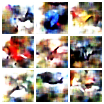}
    \caption{Denoising Attack Result}
    \label{fig:reconstructed}
\end{subfigure}
\caption{Visualization of the privacy attack on protected features. Left: original images; Middle: protected features using Layer-wise Relevance Propagation (LRP); Right: reconstructed images from the denoising attack, demonstrating potential privacy vulnerabilities.}
\label{fig:inversion_attack}
\end{figure*}
\section{Privacy Outlook and Considerations in Federated Learning}
Privacy remains a critical challenge in federated learning systems \cite{kairouz2021advances, li2020federated}. Recent investigations have revealed significant vulnerabilities to various privacy attacks, particularly model inversion attacks \cite{geiping2020inverting} and membership inference attacks \cite{nasr2019comprehensive}, which can potentially reconstruct private training data or determine if specific samples were used during training. These concerns are particularly relevant in our context, as our method introduces additional data sharing mechanisms that could potentially expand the attack surface.

Several promising approaches exist to enhance privacy guarantees in federated systems. Differential privacy (DP) \cite{dwork2014algorithmic, abadi2016deep} has emerged as a leading framework, offering mathematical guarantees for privacy preservation. While traditional DP mechanisms often involve adding calibrated noise to model updates \cite{dp-sgd}, recent advances in adaptive noise scaling \cite{zhang2022adaptive} and personalized privacy budgets \cite{wang2023personalized} provide more nuanced trade-offs between model utility and privacy protection.

To strengthen the privacy guarantees of our methodology, we propose several potential extensions:

\begin{itemize}
    \item Integration of secure aggregation protocols \cite{bonawitz2017practical} combined with local differential privacy to protect individual client contributions
    \item Implementation of gradient pruning and compression techniques \cite{lin2022privacy} to minimize potential information leakage while maintaining performance
    \item Adoption of privacy-preserving attribution mechanisms that provide interpretability without compromising sensitive information \cite{singh2020attribution}
\end{itemize}

Future work should investigate the empirical privacy-utility trade-offs of these approaches in our context, particularly focusing on how they interact with our attribution-based sampling strategy. Additionally, exploring privacy-preserving techniques specifically designed for handling interpretation mechanisms could provide valuable insights for the broader federated learning community.
\section{Extended Discussion on Attribution Methods and Sparsification}
\label{app:sparsification_discussion}

\paragraph{Motivation and Setup.}
In our main text, we evaluated how different \emph{attribution methods} perform under increasing sparsification of the input image. Specifically, we applied a thresholded binary mask $\mathbf{m}$ to retain only the top-$s\%$ most salient pixels, where $s$ ranges from 60 to 85. We observed that \textbf{FedXDS maintains stable performance} across all attribution techniques (LRP, Gradient, SmoothGrad, Integrated Gradients), despite removing a significant fraction of pixels at higher $s$ values.

These findings suggest that \textbf{selecting truly discriminative pixels} is crucial for accurate predictions and that \emph{how} those pixels are identified can strongly influence final model performance under extreme sparsification.

\subsection{Why LRP is Particularly Robust at High Sparsification}

\paragraph{Structured vs.\ Isolated Features.}
Our experiments consistently show that LRP \cite{bach2015pixel, LRP-Review, LPR-coherence} outperforms gradient-based approaches. Unlike methods such as SmoothGrad or basic gradients—which often highlight \emph{isolated} pixels deemed locally important—LRP is designed to \emph{propagate} relevance backward through each layer of the network. This layer-wise relevance propagation produces \emph{coherent, contiguous} attributions. Consequently, at high sparsity (e.g., $s=85$), the retained pixel regions form \emph{semantically consistent} patches rather than scattered points, preserving enough context to maintain robust classification performance.

\paragraph{Numerical Stability via the \texorpdfstring{$\varepsilon$}{epsilon}-Rule.}
LRP’s use of the $\varepsilon$-rule helps mitigate saturation effects, ensuring that small variations in activation do not excessively amplify or nullify the attributions. In contrast, gradient-based methods can encounter vanishing or exploding gradients, particularly in deeper networks, leading to unstable attributions. When such attributions are thresholded, the final retained pixels may be \emph{suboptimal} for preserving discriminative features. By contrast, LRP’s numerical stability and hierarchical propagation framework preserve more \emph{functionally relevant} pixels under high-threshold masking.

\paragraph{Saliency vs.\ Structural Relevance.}
While gradient methods excel at detecting certain “hotspot” pixels, they sometimes overlook broader \emph{structural} cues. By design, LRP aims to reveal how individual neurons (and layers) contribute to the network’s decision in a cumulative manner, capturing important spatial patterns. At high sparsification, \emph{having a coherent region} of relevant pixels (rather than discrete points) can maintain classification accuracy. Our results confirm that this approach is particularly advantageous in federated scenarios, where \emph{masked, privacy-preserving data} must still retain sufficient representational power for shared training.

\subsection{Implications for FedXDS}
\label{app:implications_fedxds}

Overall, these experiments confirm that the \emph{efficacy} of FedXDS in federated scenarios hinges on:

\begin{enumerate}
    \item \textbf{Effectively identifying task-relevant regions} (i.e., robust attributions), so that the most informative pixels remain under high sparsification.
    \item \textbf{Preserving} enough spatially coherent context to enable generalization despite removing most of the input.
\end{enumerate}

The superior performance of LRP underscores the importance of \emph{structural coherence} in saliency maps when only a small fraction of the image is shared. Combined with FedXDS’s privacy-preserving pipeline, LRP’s ability to highlight \emph{contiguous} and \emph{semantically meaningful} areas ensures that even heavily masked images can support effective model training.

\section{Training Porgression}
We plot the training progression for the CIFAR10 and CIFAR100 datasets of FedXDS versus some common baselines. It shows that the convergence speed and overall progression is faster and smoother for our method. 
\begin{figure*}[t]
    \centering
    \begin{subfigure}[b]{0.45\textwidth}
        \centering
        \includegraphics[width=\textwidth]{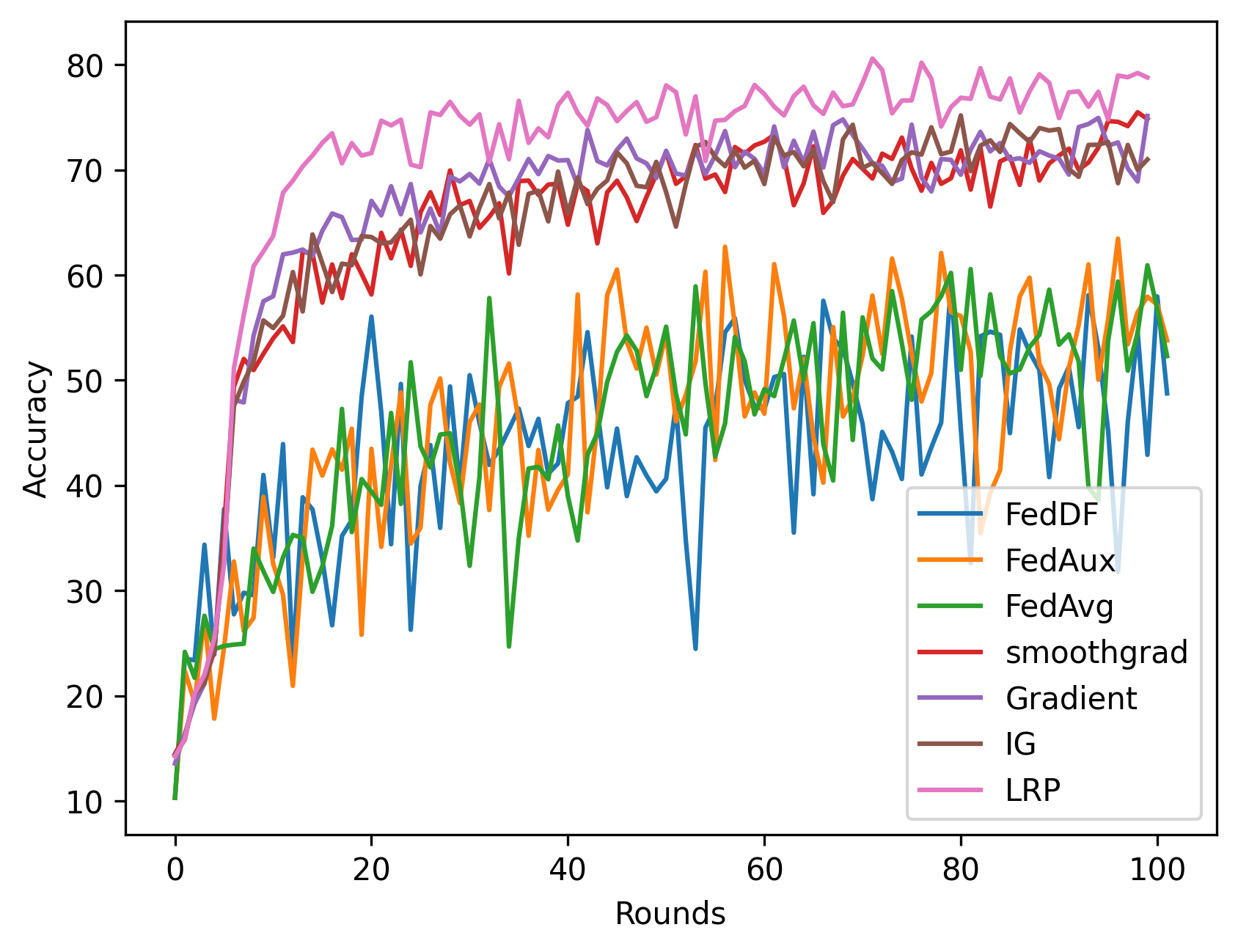}
        \caption{Training Progression of FedXDS with CIFAR10, K=10 and $\alpha=0.1$.}
        \label{fig:cifar10}
    \end{subfigure}
    \hfill
    \begin{subfigure}[b]{0.45\textwidth}
        \centering
        \includegraphics[width=\textwidth]{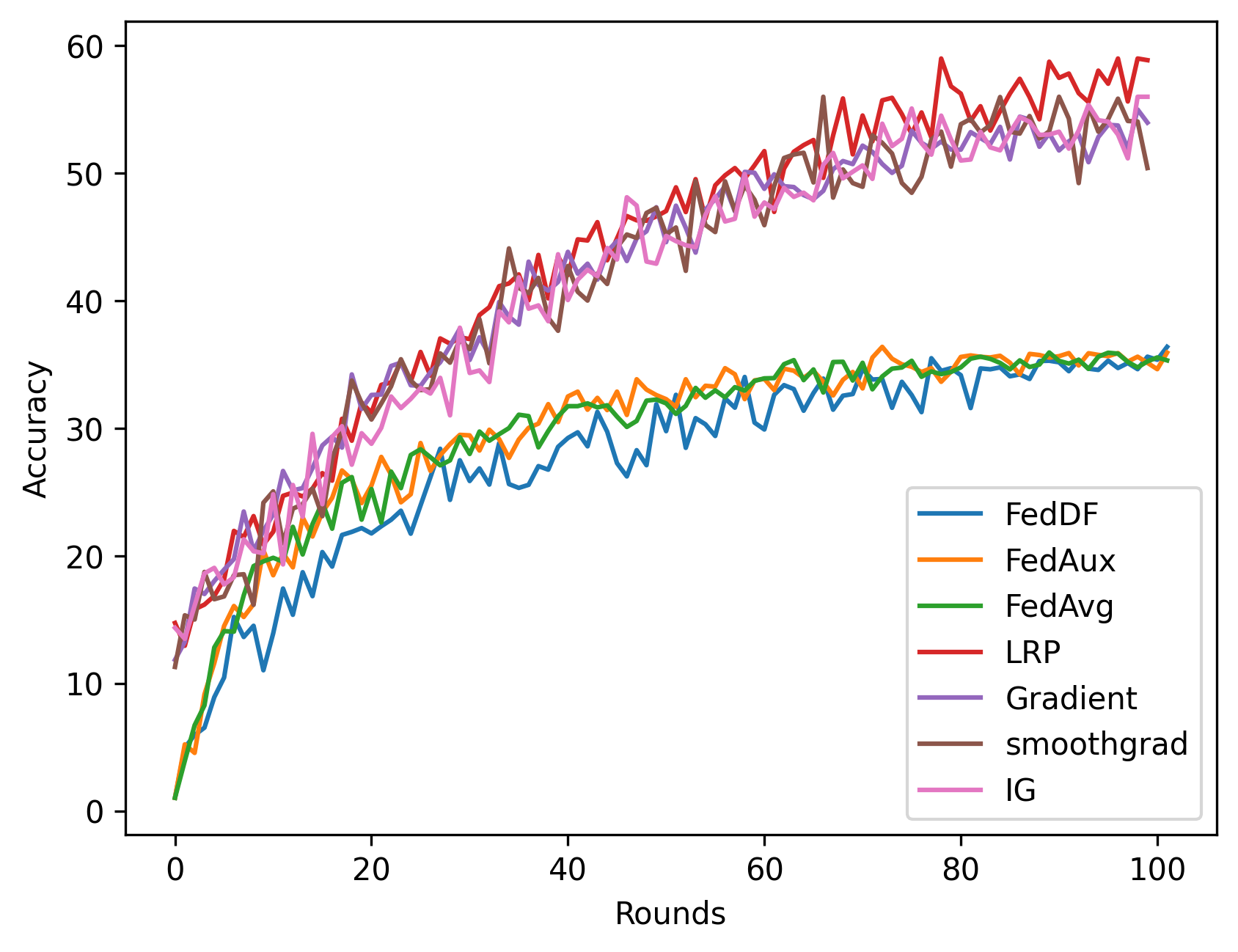}
        \caption{Training Progression of FedXDS with CIFAR100, K=10 and $\alpha=0.1$.}
        \label{fig:cifar100}
    \end{subfigure}
    \label{fig:both}
\end{figure*}
\section{Performance and Number of Samples}
We also conduct experiments on how the performance of our method depends on the number of samples shared. Specifically, we share decreasing fractions of the original dataset size of each client and track the performance. The results are shown in \autoref{tab:acc_vs_samples}:
\begin{table*}[t]
\centering
\renewcommand{\arraystretch}{1.2}
\caption{Test accuracy (\%) for different sample sizes across various attribution methods.}
\begin{tabular}{c*{4}{c}}
\toprule
Fraction of Original Samples & Gradient & SmoothGrad & Int. Gradients & LRP \\
\midrule
0.5 & 70.12 & 69.85 & 71.23 & \textbf{77.61} \\
0.6 & 71.34 & 71.02 & 72.45 & \textbf{78.91} \\
0.7 & 72.15 & 71.88 & 73.12 & \textbf{79.05} \\
0.8 & 73.45 & 73.21 & 74.56 & \textbf{80.38} \\
0.9 & 74.82 & 74.53 & 75.89 & \textbf{81.61} \\
1.0 & 76.12 & 75.89 & 77.23 & \textbf{83.67} \\
\bottomrule
\end{tabular}
\label{tab:acc_vs_samples}
\end{table*}
The table shows classification accuracy for different sample sizes from 0.5 to 1.0 across four attribution methods. LRP consistently achieves the best performance, with accuracy ranging from 77.61\% to 83.67\%, while the other methods (Gradient, SmoothGrad, and Integrated Gradients) perform 6-8\% worse but improve similarly with increasing sample size.
\section{Significance Tests on Improvements}
\begin{table}[h]
\centering
\caption{Statistical significance of FedXLRP's accuracy improvements on CelebA and FEMNIST. Table shows p-values from a one-tailed t-test comparing FedXLRP against each baseline. The reference performance for FedXLRP is $91.55 \pm 0.48$ (CelebA) and $89.03 \pm 0.35$ (FEMNIST).}
\label{tab:p-values-celeba-femnist}
\footnotesize
\setlength{\tabcolsep}{4pt}
\begin{tabular}{lcc}
\toprule
\textbf{Baseline Method} & \textbf{CelebA (p-value)} & \textbf{FEMNIST (p-value)} \\
\midrule
FedAvg    & $<0.001$ & $<0.001$ \\
FedProx   & $<0.001$ & $<0.001$ \\
SCAFFOLD  & $<0.001$ & $<0.001$ \\
FedDyn    & $<0.001$ & $<0.001$ \\
FedSAM    & $<0.001$ & $0.002$ \\
FedDISCO  & $0.001$ & $0.001$ \\
\midrule
FedFed    & $0.009$ & $0.007$ \\
FedFTG    & $0.006$ & $0.004$ \\
FedGen    & $0.002$ & $<0.001$ \\
FedAux    & $0.001$ & $<0.001$ \\
FedDF     & $<0.001$ & $<0.001$ \\
\bottomrule
\end{tabular}
\end{table}

\begin{table*}[t]
\centering
\renewcommand{\arraystretch}{1.2}
\caption{Statistical significance (p-values) of FedXLRP's accuracy improvement over other federated learning methods. P-values are derived from a one-tailed t-test for each experimental setting. "n.s." denotes a non-significant result ($p \geq 0.05$) where FedXLRP did not outperform the baseline.}
\label{tab:p-values-main}
\resizebox{\textwidth}{!}{%
\begin{tabular}{lcccccccccccc}
\toprule
Dataset & \multicolumn{4}{c}{CIFAR-10} & \multicolumn{4}{c}{CIFAR-100} & \multicolumn{4}{c}{Tiny-ImageNet} \\
\midrule
 & \multicolumn{2}{c}{K=10} & \multicolumn{2}{c}{K=100} & \multicolumn{2}{c}{K=10} & \multicolumn{2}{c}{K=100} & \multicolumn{2}{c}{K=10} & \multicolumn{2}{c}{K=100} \\
\midrule
& $\alpha=0.05$ & $\alpha=0.1$ & $\alpha=0.05$ & $\alpha=0.1$ & $\alpha=0.05$ & $\alpha=0.1$ & $\alpha=0.05$ & $\alpha=0.1$ & $\alpha=0.05$ & $\alpha=0.1$ & $\alpha=0.05$ & $\alpha=0.1$ \\
\midrule
FedAvg   & $<0.001$ & $<0.001$ & $<0.001$ & $<0.001$ & $<0.001$ & $<0.001$ & $<0.001$ & $<0.001$ & $<0.001$ & $<0.001$ & $<0.001$ & $<0.001$ \\
FedProx  & $<0.001$ & $<0.001$ & $<0.001$ & $<0.001$ & $<0.001$ & $<0.001$ & $0.002$ & $<0.001$ & $0.003$ & $<0.001$ & $<0.001$ & $0.015$ \\
FedDyn   & $<0.001$ & $0.002$ & $<0.001$ & $<0.001$ & $0.005$ & $<0.001$ & $0.001$ & $<0.001$ & $0.002$ & $0.004$ & $<0.001$ & $0.021$ \\
SCAFFOLD & $<0.001$ & $<0.001$ & $<0.001$ & $<0.001$ & $0.001$ & $<0.001$ & $<0.001$ & $<0.001$ & $<0.001$ & $0.005$ & $<0.001$ & $0.011$ \\
FedSAM   & $<0.001$ & $<0.001$ & $<0.001$ & $<0.001$ & $0.003$ & $<0.001$ & $0.001$ & $<0.001$ & $0.001$ & $0.003$ & $<0.001$ & $0.018$ \\
FedDISCO & $<0.001$ & $<0.001$ & $<0.001$ & $<0.001$ & $0.002$ & $<0.001$ & $<0.001$ & $<0.001$ & $<0.001$ & $0.002$ & $<0.001$ & $0.019$ \\
\midrule
FedFed   & $0.012$ & $0.025$ & $0.021$ & $0.018$ & n.s. & $0.011$ & $0.024$ & $0.004$ & $0.015$ & $0.009$ & $0.011$ & n.s. \\
FedFTG   & $<0.001$ & $0.003$ & $<0.001$ & $0.002$ & $<0.001$ & $0.004$ & $<0.001$ & $<0.001$ & $0.001$ & $<0.001$ & $<0.001$ & $0.025$ \\
FedGen   & $<0.001$ & $<0.001$ & $<0.001$ & $<0.001$ & $<0.001$ & $<0.001$ & $<0.001$ & $<0.001$ & $0.006$ & $0.007$ & $<0.001$ & $0.017$ \\
FedAux   & $<0.001$ & $<0.001$ & $<0.001$ & $<0.001$ & $<0.001$ & $<0.001$ & $<0.001$ & $<0.001$ & $<0.001$ & $<0.001$ & $<0.001$ & $0.013$ \\
FedDF    & $<0.001$ & $<0.001$ & $<0.001$ & $<0.001$ & $<0.001$ & $<0.001$ & $<0.001$ & $<0.001$ & $<0.001$ & $0.001$ & $<0.001$ & $0.012$ \\
\midrule
FedXIG   & $<0.001$ & $<0.001$ & $<0.001$ & $<0.001$ & $0.008$ & $0.009$ & $0.003$ & $0.001$ & $0.004$ & $0.003$ & $<0.001$ & $0.020$ \\
FedXGrad & $<0.001$ & $<0.001$ & $<0.001$ & $<0.001$ & $0.011$ & $0.005$ & $0.002$ & $0.001$ & $0.003$ & $0.001$ & $<0.001$ & $0.014$ \\
FedXSG   & $<0.001$ & $<0.001$ & $<0.001$ & $<0.001$ & $0.023$ & $0.008$ & $0.001$ & $0.001$ & $0.004$ & $0.002$ & $<0.001$ & $0.016$ \\
\bottomrule
\end{tabular}%
}
\end{table*}

To rigorously validate the performance gains of our proposed method, we conducted statistical significance tests comparing \texttt{FedXLRP} to all other baselines. We performed a one-tailed t-test for each experimental setting, with the null hypothesis being that \texttt{FedXLRP}'s performance is not superior to the baseline. As demonstrated by the consistently low p-values, the improvements achieved by \texttt{FedXLRP} are statistically significant across all datasets and heterogeneity configurations. A comprehensive summary of these p-values for our main experiments is presented in Table~\ref{tab:p-values-main} in the appendix. Similarly, the significance of our method's substantial gains on the CelebA and FEMNIST datasets is detailed in Appendix Table~\ref{tab:p-values-celeba-femnist}. These results confirm that the observed superiority of \texttt{FedXLRP} is not due to random chance but represents a consistent and meaningful advantage.

\section{Related Work}
\label{sec:appendix_related_work}

Federated Learning (FL) provides a framework for training models on decentralized data without explicit data sharing. The foundational algorithm, FedAvg \cite{FedAvg}, aggregates locally trained models to update a global model. However, its performance degrades significantly when client data is not independently and identically distributed (non-IID). Many subsequent works have sought to address this challenge.

\subsection{Client Drift Correction and Regularization}
One major line of research focuses on mitigating "client drift," where local models diverge due to heterogeneous data. FedProx \cite{FedProx} introduces a proximal term to the local objective function, restricting local updates from moving too far from the global model. Similarly, methods like SCAFFOLD \cite{SCAFFOLD} and the work of \cite{mikkelschmidt} introduce control variates to correct for the variance introduced by non-IID data, leading to improved convergence and stability. FedDyn \cite{FedDyn} addresses client heterogeneity by dynamically regularizing local objectives based on historical updates, aligning local and global optima. Other specialized approaches include FedBN \cite{FedBN}, which maintains local batch normalization parameters to handle feature shifts, FedNova \cite{FedNova}, which normalizes local updates to ensure fair client contributions, and MOON \cite{MOON}, which uses model-contrastive learning to align local and global model representations. \cite{FedSAM} attempts to regularize the loss landscape of clients, whereas \cite{Feddisco} uses a discrepancy aware approach. 

\subsection{Data Sharing and Knowledge Distillation}
Another prominent line of work aims to directly tackle data heterogeneity through various forms of data or knowledge sharing. FedDF~\cite{FedDF} aggregates knowledge from client models into a global model by using ensemble distillation on an unlabeled public dataset, unifying disparate local knowledge without sharing the private data itself. Some methods leverage generative models to create shareable synthetic data. FedGen \cite{FedGen} synthesizes class-conditional feature-space representations to align distributions between clients and the server. FedFTG \cite{FedFTG} transfers knowledge specifically at the feature level by generating pseudo-data and using it in an ensemble-distillation setup. FedFed \cite{FedFed} combines feature distillation with a variational auto-encoder to generate data under differential privacy constraints. FedAux \cite{data_sharing_FL} shares differentially private model predictions in a distillation framework, though it also requires a public dataset.

While powerful, these data-sharing approaches often introduce significant computational and communication overhead from generating and transmitting data. Furthermore, as we show in our experiments, generator-based methods may not guarantee strong empirical privacy. This highlights the central challenge that we seek to address: enabling effective knowledge sharing that mitigates heterogeneity while simultaneously preserving privacy and maintaining computational efficiency. Our work, \texttt{FedXDS}, diverges from these approaches by leveraging XAI to extract and share only the most salient feature importances—a highly compact and abstract representation that is inherently more private and efficient than sharing raw data, features, or predictions.
\section{Warmup and Computational Efficiency}

\subsection{Warmup Rounds}
We investigate the effect of varying the number of warmup rounds—used to initialize the model before applying attribution-based feature selection—on downstream model performance. Warmup is performed using standard FedAvg for $R_{\text{warmup}} \in \{0, 5, 10, 15\}$ rounds. The results, demonstrate a consistent improvement in test accuracy with increasing warmup. Across all datasets (CIFAR-10, CIFAR-100, Tiny-ImageNet), performance improves sharply from 0 to 10 rounds and then begins to plateau.

This behavior aligns with the need for stable and meaningful attributions: early in training, the model has not yet learned reliable patterns, and attribution maps may reflect noise or spurious correlations. A few rounds of pretraining help the model to focus on task-relevant features, producing more semantically meaningful relevance maps for masking. Beyond 10 rounds, marginal accuracy gains diminish, suggesting that longer warmup adds little benefit but incurs extra communication overhead. Thus, we fix $R_{\text{warmup}}=10$ as a good trade-off between accuracy and efficiency in our main experiments.

\subsection{Computational Efficiency}
FedXDS is designed with federated settings in mind, where client devices often have limited compute and memory. Our approach offers key efficiency advantages over prior methods that rely on data generation:

\begin{itemize}
  \item \textbf{One-time attribution pass:} Each client computes attribution maps using a single backward pass through a pretrained (warmup) model for each sample. This step is comparable in cost to one training epoch and is performed only once.
  \item \textbf{No generators or decoders:} Unlike FedGen and FedFTG, which require training and updating GANs or feature generators at every communication round, FedXDS relies on simple masking and additive noise. This eliminates the need for high-dimensional generation pipelines and avoids mode collapse or instability issues.
  \item \textbf{No latent encoders:} FedFed uses variational autoencoders (VAEs) for feature distillation, which requires both encoder and decoder networks, often doubling the model size and computation. Moreover, VAEs require tuning of KL regularization terms and are known to be sensitive to data heterogeneity.
\end{itemize}

In contrast, FedXDS operates with minimal model modifications and negligible per-round overhead. Once attribution-based masks are computed and privacy noise is added, the resulting masked dataset can be reused throughout training. This design not only reduces client-side resource requirements but also improves communication efficiency by accelerating convergence (as shown in Table~2 of the main paper).

We examine computation times (seconds per round) for CIFAR-10 (\(\alpha=0.1\), 10 clients, 50\% participation, 100 rounds), in  for generator training are: FedFed (79), FedGen (34), FedFTG (37), and FedXDS (32). FedFed requires 15 rounds to train the generator, while FedGen and FedFTG update every round. Our method requires a single backward pass per sample one time only, limiting computation to a single round. 

In summary, FedXDS strikes a favorable balance between computational cost, model performance, and privacy—making it particularly well-suited for real-world federated learning deployments where resources are constrained.

\end{document}